\documentclass[sigconf]{acmart}

\usepackage{tikz,pifont,algorithm,algorithmic,tabularx,multirow,amsmath,comment,threeparttable,booktabs,adjustbox,caption,xcolor}
\usepackage{hyperref}
\usepackage[T1]{fontenc}
\definecolor{mygreen}{rgb}{0,0.7,0}

\copyrightyear{2024}
\acmYear{2024}
\setcopyright{acmlicensed}
\acmConference[CCS '24] {Proceedings of the 2024 ACM SIGSAC Conference on Computer and Communications Security}{October 14--18, 2024}{Salt Lake City, UT, USA.}
\acmBooktitle{Proceedings of the 2024 ACM SIGSAC Conference on Computer and Communications Security (CCS '24), October 14--18, 2024, Salt Lake City, UT, USA}
\acmISBN{979-8-4007-0636-3/24/10}
\acmDOI{10.1145/3658644.3670293}

\begin{document}

\title{\emph{SUB-PLAY}: Adversarial Policies against Partially Observed Multi-Agent Reinforcement Learning Systems}

\author{Oubo Ma}
\affiliation{
    \institution{Zhejiang University}
    \country{}}
\email{mob@zju.edu.cn}
\orcid{0000-0002-6572-972X}

\author{Yuwen Pu}
\authornote{Yuwen Pu and Shouling Ji are the co-corresponding authors.}
\affiliation{
    \institution{Zhejiang University}
    \country{}}
\email{yw.pu@zju.edu.cn}

\author{Linkang Du}
\affiliation{
    \institution{Xi'an Jiaotong University}
    \country{}}
\email{linkangd@gmail.com}

\author{Yang Dai}
\affiliation{
    \institution{Laboratory for Big Data and Decision}
    \country{}}
\email{daiyang2000@163.com}

\author{Ruo Wang}
\affiliation{
    \institution{Chinese Aeronautical Establishment}
    \country{}}
\email{kurt_ashtray@163.com}

\author{Xiaolei Liu}
\affiliation{
    \institution{Institute of Computer Application, China Academy of Engineering Physics}
    \country{}}
\email{luxaole@gmail.com}

\author{Yingcai Wu}
\affiliation{
    \institution{Zhejiang University}
    \country{}}
\email{ycwu@zju.edu.cn}

\author{Shouling Ji}
\authornotemark[1]
\affiliation{
    \institution{Zhejiang University}
    \country{}}
\email{sji@zju.edu.cn}

\begin{abstract}
Recent advancements in multi-agent reinforcement learning (MARL) have opened up vast application prospects, such as swarm control of drones, collaborative manipulation by robotic arms, and multi-target encirclement.
However, potential security threats during the MARL deployment need more attention and thorough investigation.
Recent research reveals that attackers can rapidly exploit the victim's vulnerabilities, generating adversarial policies that result in the failure of specific tasks.
For instance, reducing the winning rate of a superhuman-level Go AI to around 20\%.
Existing studies predominantly focus on two-player competitive environments, assuming attackers possess complete global state observation.

In this study, we unveil, for the first time, the capability of attackers to generate adversarial policies even when restricted to partial observations of the victims in multi-agent competitive environments.
Specifically, we propose a novel black-box attack (\emph{SUB-PLAY}) that incorporates the concept of constructing multiple subgames to mitigate the impact of partial observability and suggests sharing transitions among subpolicies to improve attackers' exploitative ability.
Extensive evaluations demonstrate the effectiveness of \emph{SUB-PLAY} under three typical partial observability limitations.
Visualization results indicate that adversarial policies induce significantly different activations of the victims' policy networks.
Furthermore, we evaluate three potential defenses aimed at exploring ways to mitigate security threats posed by adversarial policies, providing constructive recommendations for deploying MARL in competitive environments.
\end{abstract}



\maketitle

\section{Introduction}

Multi-agent reinforcement learning (MARL) has succeeded remarkably in diverse domains, from StarCraft II~\cite{rashid2020monotonic} to cyber-physical systems~\cite{wang2016towards}, strategic maneuvers~\cite{fernandez2021multi}, and social science~\cite{leibo2017multi}.
Currently, MARL predominantly emphasizes improving algorithm performance across various tasks, yet there is a noticeable lack of consideration for security aspects.

\begin{figure}
\centering
\includegraphics[height=1.3 in]{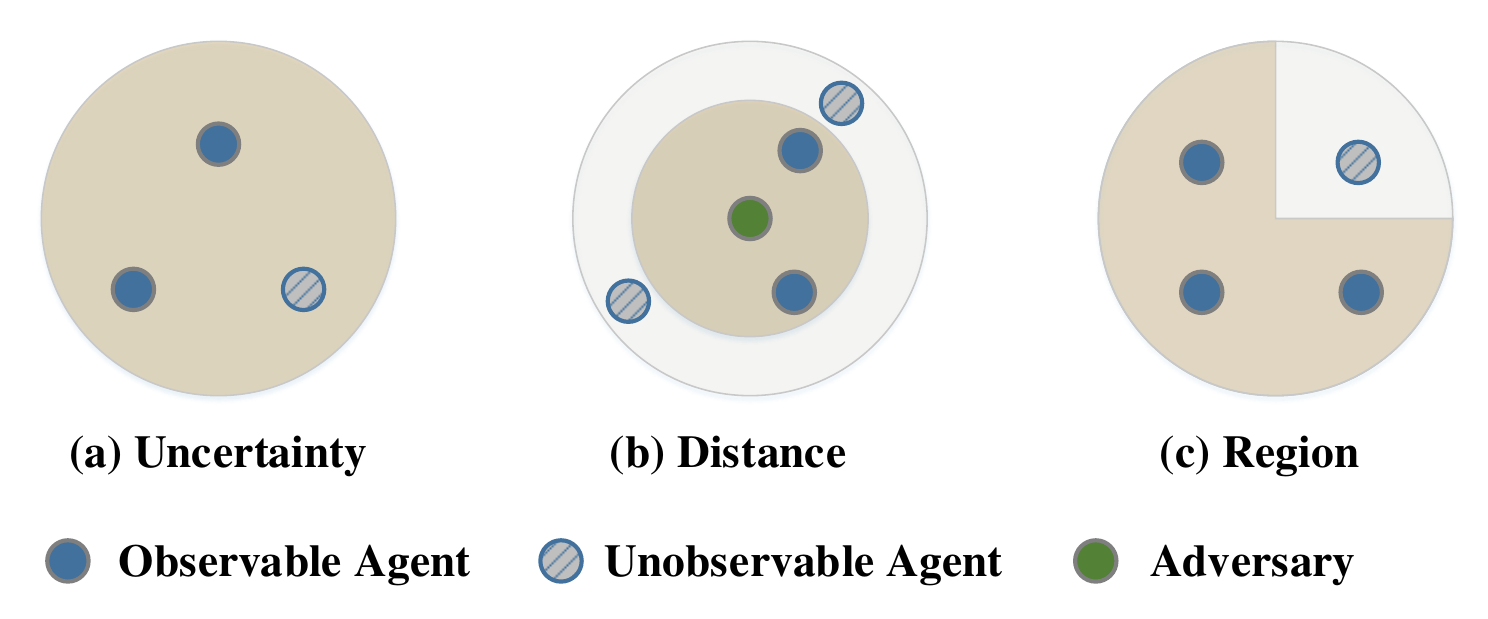}
\caption{Three partially observable limitations in multi-agent environments.}
\label{fig:limitation}
\end{figure}

Recent research~\cite{wang2023adversarial} has unveiled that even state-of-the-art reinforcement learning (RL) policies exhibit weaknesses and vulnerabilities in competitive environments.
Therefore, an attacker can employ adversarial policies to induce the victim's policies to trigger vulnerabilities, resulting in a significant performance decline, possibly even leading to a loss of sequential decision-making capability.

For instance, \emph{Victim-play}~\cite{gleave2020adversarial} is a black-box attack framework designed for adversarial policy generation, where the attacker can interact with the victim without requiring access to the victim's policy or environmental perturbations.
However, \emph{Victim-play} is designed for two-player competitions, where the victim operates as a single-agent system, such as a superhuman-level Go AI~\cite{wang2023adversarial}.
It encounters challenges when striving to sustain a stable attack performance within multi-agent  competitions, with the victim functioning as a multi-agent system (MAS), for instance, a drone swarm~\cite{zhou2022swarm}.
The widespread prevalence of partial observability exacerbates this challenge, as attackers are unable to access complete global state information.
This may result in adversarial strategies being unable to converge due to fluctuations or getting trapped in poorly performing local optima.

Partial observability is primarily attributed to three limitations (see~\autoref{fig:limitation}):
(1) Uncertainty Limitation: This occurs when partial observability arises due to the constraints imposed by unpredictable environmental events. Examples include obstacle occlusion, noisy measurements, and sensor anomalies.
(2) Distance Limitation: This refers to situations where the relative distance between agents exceeds their perceptual range, determined by the sensors deployed by MASs, such as LiDARs and millimeter-wave radars~\cite{ren2019security}.
(3) Region Limitation: Incomplete observations result from privacy concerns, security constraints, or rule restrictions, where specific boundaries define the region and could represent geographical areas or logical ranges. Examples in this regard encompass restricted areas due to permission controls or competitions with incomplete information, such as financial markets or Texas Hold'em poker.

This paper introduces \emph{SUB-PLAY}, a novel black-box attack framework aimed at adversarial policy generation in partially observed multi-agent competitive environments.
Our intuition lies in the divide-and-conquer principle, decomposing the attack into multiple subgames.
Each subgame is then modeled as a partially observable stochastic game (POSG)~\cite{littman1994markov}, and MARL is employed to solve and obtain the corresponding subpolicy.
Finally, we integrate all subpolicies in a hard-coded format to generate the ultimate adversarial policy.
Our main challenge is the ineffectiveness of attacks caused by data imbalance.
Specifically, the attacker records the interactions at each time step in the form of transitions and allocates these transitions to a specific subgame replay buffer based on the observed number of victim agents.
However, the number of transitions in each buffer is uneven due to varying probabilities of each subgame occurrence.
This imbalance may lead to undertraining of some subpolicies.
To mitigate this issue, we propose a transition dissemination mechanism that facilitates the sharing of transition from proximity subgames.

Extensive evaluation results demonstrate that \emph{SUB-PLAY} can effectively address the aforementioned three partial observability limitations and outperform \emph{Victim-play} in Predator-prey and World Communication, two representative multi-agent competitive environments open-sourced by OpenAI~\cite{lowe2017multi}.
Compared to normal opponents, t-SNE analysis reveals a significant difference in the activations of the victim's policy network during interactions with adversarial policies.
The scalability evaluation indicates that attackers can adjust the granularity of subgame construction to expand the applicability of our method.
Moreover, \emph{SUB-PLAY} is algorithm-agnostic, \emph{i.e.}, applicable to both distributed and centralized MARL algorithms.

To explore strategies for mitigating adversarial policies, we evaluate three potential defenses.
The results indicate that adversarial retraining is insufficient to counteract adversarial policies, while policy ensemble and fine-tuning could only moderately reduce the effectiveness of attacks.
Nevertheless, insights from the evaluation suggest that defenders may mitigate security risks posed by adversarial policies through flexible deployment techniques.
For example, periodically updating policies or increasing the diversity of policies in policy ensemble.

In summary, the paper makes the following contributions:

\begin{itemize}
  \item To the best of our knowledge, \emph{SUB-PLAY}\footnote{The source code of \emph{SUB-PLAY} can be found at \href{https://github.com/maoubo/SUB-PLAY}{https://github.com/maoubo/SUB-PLAY}.} is the first work to investigate the security threats of adversarial policies in multi-agent competitive environments, revealing that attackers can exploit vulnerabilities in the victim's policy even with partial observations.
  \item We summarize three partially observable limitations and propose an observable-driven subgame construction method to accommodate these limitations.
  \item We conduct a systematic evaluation, demonstrating that \emph{SUB-PLAY} outperforms the state-of-the-art attack framework in partially observable multi-agent competitive environments.
  \item We explore potential defenses, emphasizing that practitioners in MARL should not only focus on improving algorithm performance but also pay attention to deployment details, which is crucial in mitigating security threats posed by adversarial policies.
\end{itemize}

\section{Background}
\label{sec:background}

\subsection{Multi-Agent RL}

MARL refers to scenarios where multiple agents are involved in sequential decision-making, and their policies are updated concurrently.
This results in a non-stationary environment where the optimal policy for each agent changes over time, making the Markov property invalid~\cite{sutton2018reinforcement}.

\noindent \textbf{MARL Tasks.}
Based on the cooperation patterns among agents, MARL tasks are primarily divided into four categories:
(1) Fully Cooperative MARL, where agents typically share a common reward function and collaborate to achieve a shared objective. Examples include multi-agent pathfinding and traffic management.
(2) Fully Competitive MARL, where agents compete individually to outperform each other and pursue their objectives. These tasks are often modeled as two-player zero-sum Markov games, where cooperation between agents is impossible. Examples include Go or arm wrestling.
(3) Self-Interested MARL, where agents prioritize their benefits without considering others, as observed in domains like autonomous driving and stock trading.
(4) Mixed MARL involves a blend of cooperative and competitive behavior. In most scenarios, two competing MASs exist, but agents within the same MAS collaborate. Examples include military exercises, multi-target encirclement, and team sports. The competitive environment discussed in this paper falls within a typical class of mixed MARL tasks.

\begin{figure}
\centering
\includegraphics[height=1.55 in]{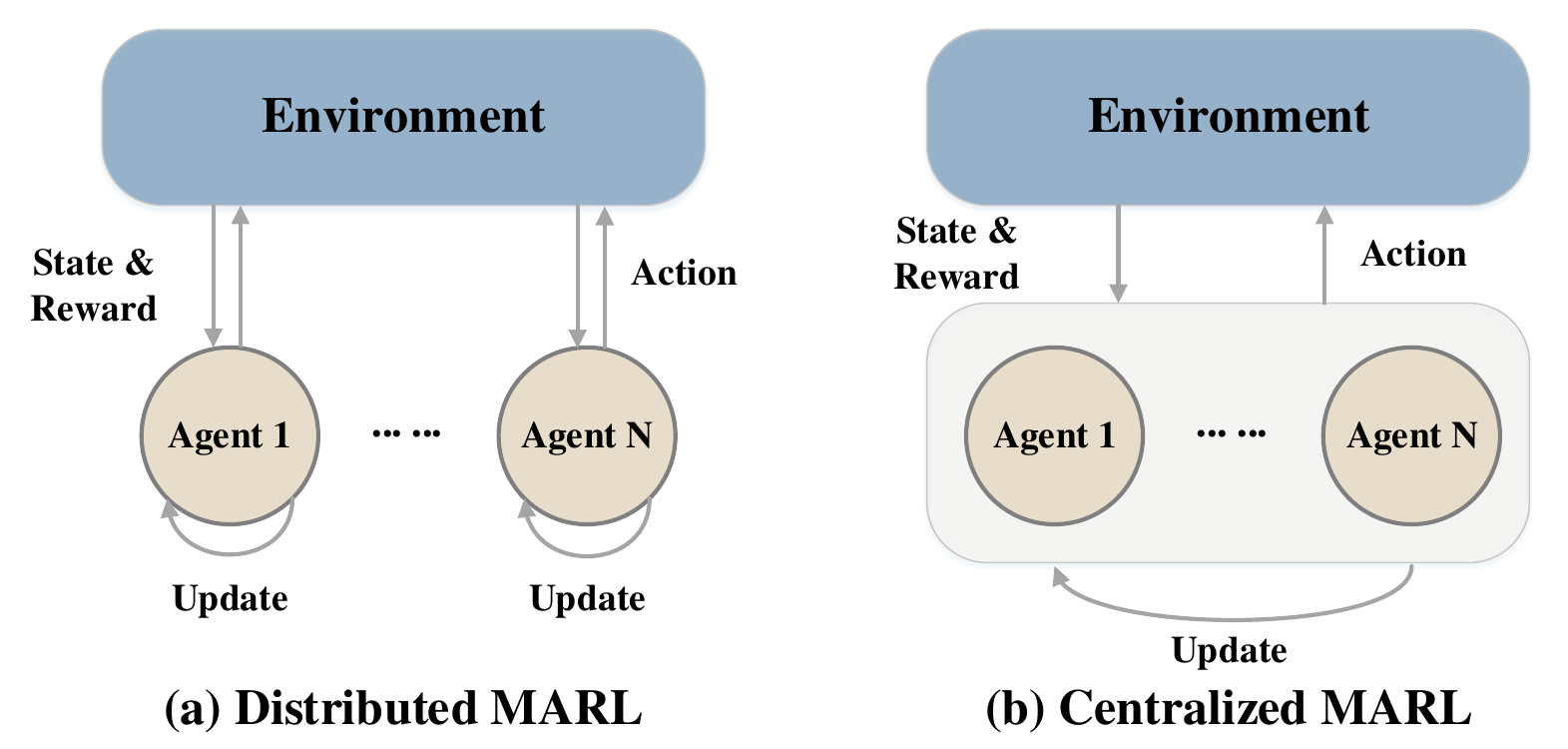}
\caption{MARL training paradigms.}
\label{fig:MARL}
\end{figure}

\noindent \textbf{Training Paradigm.}
MARL has two training paradigms based on the presence or absence of a central decision-maker~\cite{gronauer2022multi}.
(1) Distributed MARL: The algorithms assume agents update policies independently, similar to single-agent implementations.
\emph{Distributed Training Decentralized Execution (DTDE)}, a typical paradigm (\emph{e.g.}, independent Q-learning), allows efficient training and deployment without communication constraints. However, it may not be suitable for complex environments with many agents, as non-stationarity is neglected.
(2) Centralized MARL: \emph{Centralized Training Centralized Execution (CTCE)} is a centralized MARL paradigm with a centralized decision maker in training and deployment.
It achieves theoretically optimal performance but needs communication guarantees and suffers from the curse of dimensionality~\cite{nguyen2020deep}.
In contrast, \emph{Centralized Training Decentralized Execution (CTDE)} algorithms, such as QMIX, MADDPG, and MAPPO, guide agents during training but enable independent decision-making during deployment without additional communication, offering state-of-the-art performance.
\emph{SUB-PLAY} is not limited by the attacker's communication capabilities or the number of agents under its control. Therefore, \emph{SUB-PLAY} applies to both distributed and centralized MARL algorithms.

\subsection{Adversarial Policy}
\label{sec:Adversarial_Policy}

The adversarial policy is a form of action manipulation attack in which the attacker induces the black-box victim to make suboptimal decisions by controlling the actions of the adversary agents.
The training of adversarial policies relies on the competitive relationship between the attacker and the victim (typically zero-sum games), so the attacker only needs to maximize the adversary agents's reward to autonomously discover and exploit weaknesses and vulnerabilities in the RL policies deployed by the victims.

Existing research~\cite{gleave2020adversarial, wu2021adversarial, guo2021adversarial, wang2023adversarial, liu2023rethinking} predominantly focuses on two-player competitions, assuming an attacker has the privilege to interact with a victim, can obtain a complete observation of the environment at each time step, and the victim fixedly deploys a well-trained policy.
The fundamental reason for the existence of adversarial policies stems from RL's adoption of \emph{Self-play} for policy training in competitive environments~\cite{heinrich2015fictitious}. However, \emph{Self-play} cannot guarantee to reach a Nash equilibrium within finite training.
In game theory, non-equilibrium policies are inevitably exploitable.
Guided by this intuition, an attacker can manipulate its policy during training, updating it in a direction that maximizes the exploitation of the victim's vulnerabilities.

For instance, while AlphaGo-style AIs outperform human champions, adversarial policies specifically trained against them still achieve a success rate of over 77\%~\cite{wang2023adversarial}.
Remarkably, these adversarial policies fall short when facing ordinary Go enthusiasts.
This indicates that adversarial policies are highly targeted, sacrificing generalizability to intensify exploitation against a specific victim.
Therefore, an adversarial policy is a complement to RL or a figuration of its weaknesses rather than a substitute.

Finding or approximating a Nash equilibrium in a multi-agent competition is at least as hard as PPAD-complete~\cite{brown2020equilibrium}. This implies that in real deployment scenarios, MARL policies are exploitable. Therefore, MARL must be attentive to the potential risks posed by adversarial policies.

More details about the adversarial policy, including a more nuanced explanation of its existence and the upper limit of attack performance, can be found in~\autoref{app:Exploitable Ceiling}.

\section{Threat Model and Problem Formulation}
\label{sec:formulation}

\subsection{Threat Model}
\label{sec:Threat_Model}

In this paper, we propose an adversarial policy attack for mixed MARL tasks in two-team competitive environments\footnote{"Two-team competitive environments" is a subset of "multi-agent competitive environments". For clarity and readability, we use the first term in sections related to problem formulation and scheme design (\autoref{sec:formulation} and \autoref{sec:methodlogy}) and the second term in all other sections.}.

\noindent \textbf{Definition 1.}
\textit{A two-team competitive environment involves two MASs, Adversary and Victim, which consist of two sets of agents, $\mathcal{M}$ and $\mathcal{N}$, where $|\mathcal{M}|=$ M and $|\mathcal{N}|=$ N. Adversary and Victim are in full competition, while the agents within each MAS collaborate.}

\noindent \textbf{Attacker's Goal.}
Maximizing the reduction of $Victim$'s performance on a specific MARL task.

\noindent \textbf{Attacker's Capabilities.}
The attacker possesses complete control over the $Adversary$ and can update its MARL policy.
Additionally, the attacker has interaction privileges with the $Victim$, obtaining partial observations about the environment at each time step.
The attacker knows that there are $N$ agents in the $Victim$.
Apart from this, $Victim$ is regarded as a black box by the attacker. 
Moreover, the attacker is restricted from manipulating the environment.

We assume that $Victim$'s MARL policy is fixed, \emph{i.e.}, the parameters during the deployment phase are frozen (subsequent evaluations indicate that this assumption can be relaxed).
This is common for the deployment of RL on physical entities.
For instance, a manufacturer releases an encirclement MAS consisting of multiple-legged robots~\cite{fu2023deep} or drones~\cite{kaufmann2023champion}.
These physical entities fixedly deploy policies to avoid unacceptable losses due to exploration, such as robot malfunctions or drone crashes.
Even if the manufacturer offers periodic policy upgrade services, the parameters of the MAS remain fixed between two consecutive updates, and this interval might span several months or even years.
In such scenarios, the manufacturer is a potential victim.
Once the attacker generates an adversarial policy, it implies that all users deploying the MAS from that manufacturer are exposed to the threat.

\begin{figure}
\centering
\includegraphics[height=1.6 in]{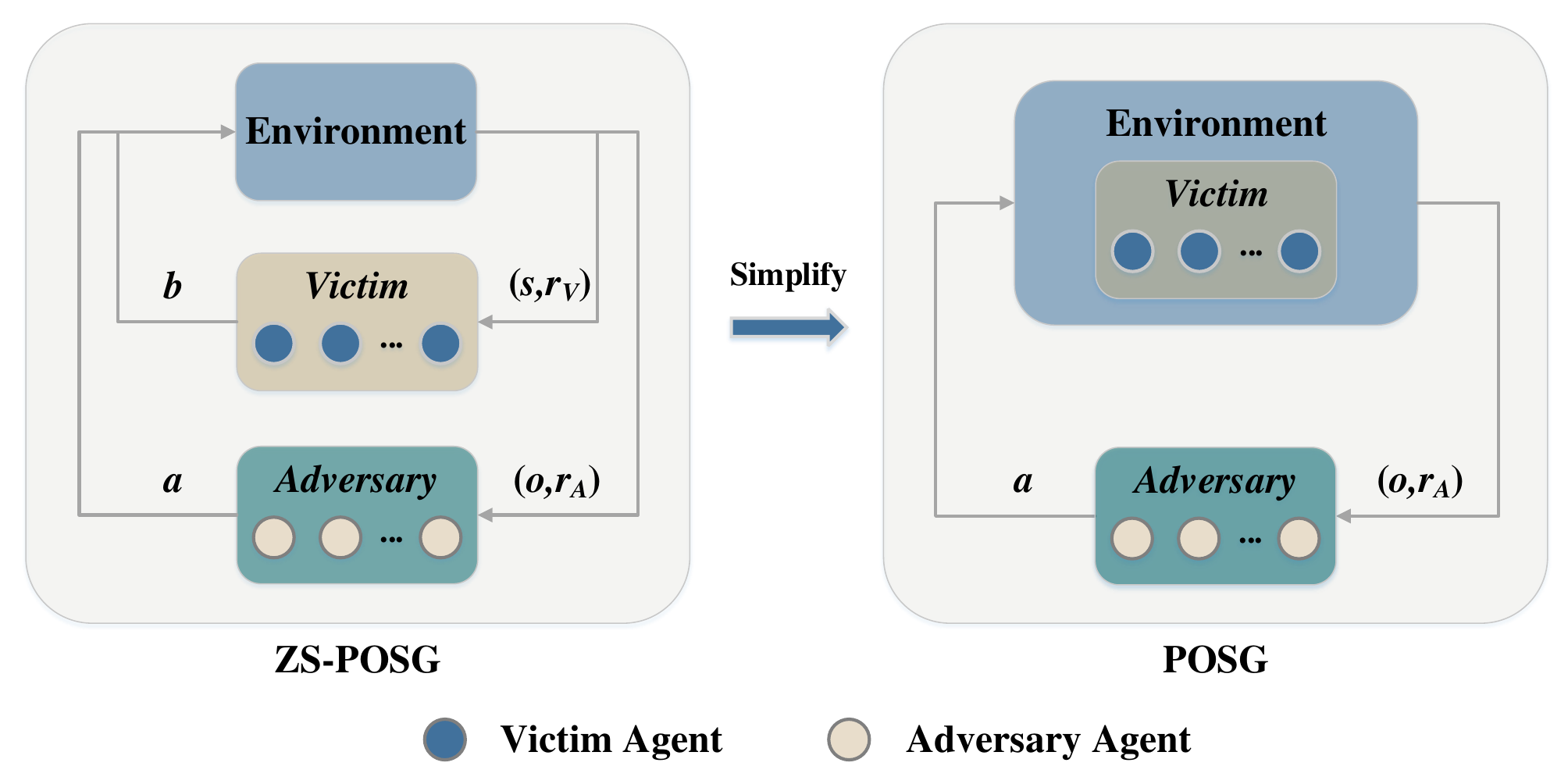}
\caption{A two-team competitive environment can be simplified from a ZS-POSG to a POSG if the joint policy of $Victim$ is fixed.}
\label{fig:model}
\end{figure}

\subsection{Problem Formulation}

Based on the threat model, we formulate the attack as a zero-sum partially observable stochastic game (ZS-POSG)~\cite{littman1994markov, zhang2021finite, hansen2004dynamic}.

\noindent \textbf{Definition 2.}
\textit{A zero-sum partially observable stochastic game is defined as}
\begin{equation}
\mathcal{G} = (\mathcal{S}, \{ \mathcal{A}^i \}, \{ \mathcal{B}^j \}, \{ \Omega^i \}, \{ \mathcal{O}^i \}, \mathcal{P}, \{ \mathcal{R}_A^i \}, \{ \mathcal{R}_V^j \}, \gamma),
\end{equation}
\textit{where $\mathcal{S}$ denotes the state space, $\mathcal{A}^i$ (resp. $\mathcal{B}^j$) denotes the action space for agent $i \in \mathcal{M}$ (resp. $j \in \mathcal{N}$). $\mathcal{A} = \prod_{i \in \mathcal{M}}\mathcal{A}^i$ and $\mathcal{B} = \prod_{j \in \mathcal{N}}\mathcal{B}^j$ denote the joint action spaces for the adversary and the victim. Each agent $i \in \mathcal{M}$ receives an observation $o^i \in \Omega^i$, and the observation function $\mathcal{O}^i:\mathcal{S} \times \mathcal{A} \times \mathcal{B} \rightarrow \Delta (\Omega ^i)$ is a probability distribution over possible subsequent observations given the previous state and the actions of all agents. Each agent $j \in \mathcal{N}$ receives the global state $s \in \mathcal{S}$. $\mathcal{P}:\mathcal{S} \times \mathcal{A} \times \mathcal{B} \rightarrow \Delta (\mathcal{S})$ represents the probability that taking joint action $a \in \mathcal{A}$ and $b \in \mathcal{B}$ in state $s \in \mathcal{S}$ results in a transition to $s' \in \mathcal{S}$. $\mathcal{R}_A^{i}:\mathcal{S} \times \mathcal{A} \times \mathcal{B} \rightarrow \mathbb{R}$ (resp. $\mathcal{R}_V^j:\mathcal{S} \times \mathcal{A} \times \mathcal{B} \rightarrow \mathbb{R}$) denotes the reward function for agent $i \in \mathcal{M}$ (resp. $j \in \mathcal{N}$). $\gamma \in [0,1)$ is the discount factor.}

Let $\pi_A \in \mathbb{P}_A: \Omega \rightarrow \Delta(\mathcal{A})$ and $\pi_V \in \mathbb{P}_V: \mathcal{S} \rightarrow \Delta(\mathcal{B})$ be the joint policies of $Adversary$ and $Victim$, respectively. $\mathbb{P}_A$ and $\mathbb{P}_V$ are their corresponding joint policy spaces.

\begin{figure*}
\centering
\includegraphics[height=3 in]{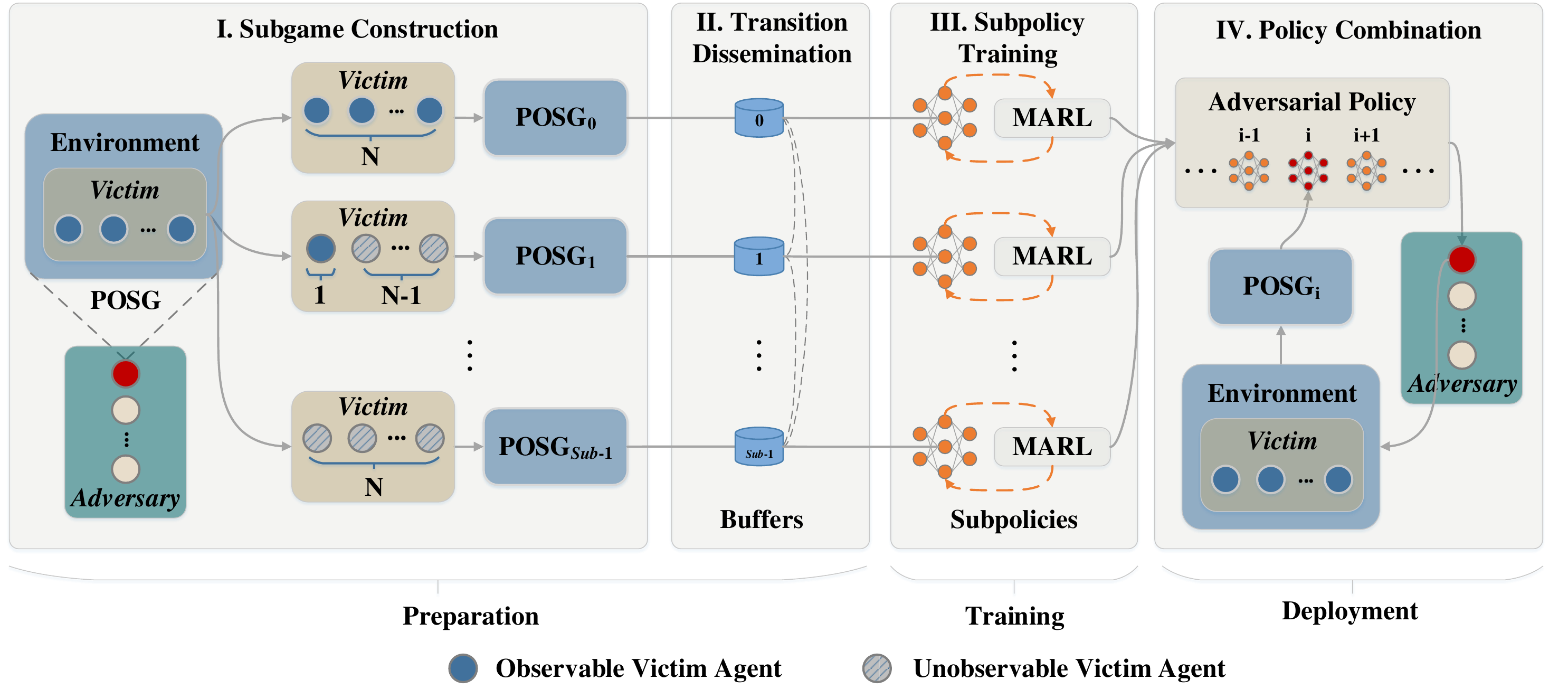}
\caption{The framework of \emph{SUB-PLAY}.}
\label{fig:architecture}
\end{figure*}

\subsection{Problem Simplification}

Inspired by~\cite{guo2021adversarial}, we define the following proposition (the proof is in \autoref{app:a}).

\noindent \textbf{Proposition 1.}
\textit{In a zero-sum partially observable stochastic game, if the victim keeps a fixed joint policy $\pi_V$, the state transition of the environment is solely dependent on the adversary's joint policy $\pi_A$.}

According to Proposition 1, the attacker can treat $Victim$ as part of the environment (as shown in \autoref{fig:model}) and simplify the attack from a ZS-POSG to a POSG:

\begin{equation}
\mathcal{G}_\alpha = (\mathcal{S}, \{ \mathcal{A}^i \}, \{ \Omega^i \}, \{ \mathcal{O}_\alpha^i \}, \mathcal{P}_\alpha, \{ \mathcal{R}_\alpha^i \}, \gamma),
\label{eq:4}
\end{equation}
retains the same state space, action space for $Adversary$, observation space, and discount factor as the original $\mathcal{G}$.
However, the observation function, transition function, and reward function for $Adversary$ are reconstructed as
\begin{equation}
\begin{aligned}
&\mathcal{O}_\alpha^i (s, a) = \mathcal{O}^i (s, a, b), \\
&\mathcal{P}_\alpha (s, a) = \mathcal{P} (s, a, b), \\
&\mathcal{R}_\alpha^i (s, a) = \mathcal{R}_A^i (s, a, b),
\end{aligned}
\end{equation}
where $a$ and $b$ are the joint actions of $Adversary$ and $Victim$, respectively.
Eventually, the attacker's objective translates into finding a policy $\pi_A \in \mathbb{P}_A$ that maximize the accumulated rewards $\sum_i^M \sum_{t=0}^T \gamma^t \mathcal{R}_\alpha^i(s_t, a_t)$ of $Adversary$, where $T$ is the time horizon.

\section{Methodology}
\label{sec:methodlogy}

This section first introduces the framework of \emph{SUB-PLAY} and then describes the design details of each step.
The intuition behind \emph{SUB-PLAY} is to adopt a divide-and-conquer strategy, decomposing a complex POSG, as depicted in \autoref{eq:4}, into multiple relatively simpler POSGs.
By tackling these simplified subgames individually, it becomes possible to address the overall complexity of the original POSG more efficiently.

\subsection{Attack Framework}

\emph{SUB-PLAY} consists of four main steps: subgame construction, transition dissemination, subpolicy training, and policy combination (see \autoref{fig:architecture}).

In the preparation phase, the attacker constructs multiple subgames based on the potentially observed number of victim agents and models each subgame as a POSG.
Each subgame initializes its own subpolicy and replay buffers.
For specific details of subgame construction, please refer to~\autoref{sec:Subgame Construction}.

To mitigate the undertraining subpolicies caused by limited interaction transition, the attacker employs a uniform transition dissemination mechanism for all agents in $Adversary$.
Then, each agent predefines a transition dissemination table, which is utilized to determine the probability of sharing each transition data among the replay buffers.
For specific details of transition dissemination, please refer to~\autoref{sec:Transition Dissemination}.

In the training phase, $Adversary$ and $Victim$ interact within the environment.
The generated transitions are stored in each replay buffer based on the probabilities determined in the transition dissemination table.
When a replay buffer accumulates a batch of transitions, the MARL algorithm updates the corresponding subpolicy.
The reward of $Adversary$ is negatively correlated with the performance of $Victim$ in the competition.
Therefore, in accordance with the MARL paradigm, each subpolicy tends to minimize the performance of $Victim$ to achieve the attack objective.
This reward-oriented process does not require any additional knowledge or human intervention.
For specific details of subpolicy training, please refer to~\autoref{sec:Subpolicy Training}.

In the deployment phase, the attacker combines subpolicies to form an adversarial policy.
Since the attacker has complete control over the adversary and stealthiness is not a concern, we implement the policy combination in a hard-coded manner.
When launching an attack, $Adversary$ determines which subgame the current competition belongs to and then switches to the corresponding subpolicy to make decisions.
More details of policy combination can be referred to~\autoref{sec:Policy Combination}.

\subsection{Subgame Construction}
\label{sec:Subgame Construction}

For a partially observed MAS, the attacker constructs subgames based on the observed agents. We define
\begin{equation}
\emph{Sub} = \emph{N} + 1,
\label{eq:sub}
\end{equation}
where \emph{Sub} indicates the number of constructed subgames and $N$ is the number of agents belonging to $Victim$.
All subgames form a set $\{ \mathcal{G}_{\alpha_k} \}_{k \in \mathcal{K}}$. $\mathcal{K}$ = $\{ 0, 1, ..., \emph{Sub} - 1 \}$ and each subgame is fomulated as a POSG:
\begin{equation}
\mathcal{G}_{\alpha_k} = (\mathcal{S}, \{ \mathcal{A}^i \}, \{ \Omega_k^i \}, \{ \mathcal{O}_{\alpha}^i \}, \mathcal{P}_\alpha, \{ \mathcal{R}_\alpha^i \}, \gamma),
\end{equation}
where the only difference in this equation from~\autoref{eq:4} is the term $\Omega_k^i \subseteq \Omega^i$. For example, if $Victim$ consists of two agents, the attacker constructs three subgames $\{ \mathcal{G}_{\alpha_0}, \mathcal{G}_{\alpha_1}, \mathcal{G}_{\alpha_2} \}$, corresponding to the cases where the attacker observes 0, 1, and 2 agents from $Victim$, respectively.

\noindent \textbf{Remark 1.}
In real-world environments, the partial information about specific agent components may be available to an attacker due to limited perspective. We propose that a conservative strategy can be adopted in high exploration cost environments, treating these agents as unobservable, while a more aggressive strategy can be employed in low exploration cost environments by treating them as observable.

\noindent \textbf{Remark 2.}
Subgame construction is scalable. In scenarios involving more agents, the attacker can choose a coarser granularity to determine the scope and number of subgames. For instance, when the victim has eight agents, attackers can construct three subgames $\{ \mathcal{G}_{\alpha_{0-2}}, \mathcal{G}_{\alpha_{3-5}}, \mathcal{G}_{\alpha_{6-8}} \}$ instead of nine.
Furthermore, the attacker could construct subgames based on regions or apply our method to scenarios where the victim is a single-agent system. In the latter case, subgames could be constructed based on the observability of different components of the single agent.

\begin{figure*}
\centering
\includegraphics[height=1.45 in]{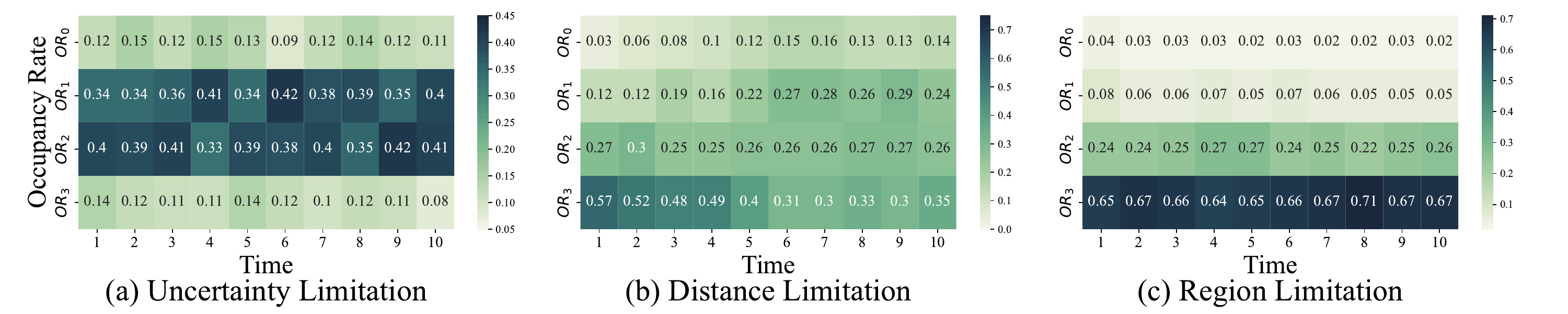}
\caption{The occupancy rate of subgames under three different limitations. The environment is Predator-prey, with $Victim$ consisting of three agents. The vertical coordinate contains the occupancy rate of four subgames $\{ \mathcal{G}_{\alpha_0}, \mathcal{G}_{\alpha_1}, \mathcal{G}_{\alpha_2} , \mathcal{G}_{\alpha_3} \}$. As time progresses, the attacker's policy will be updated.}
\label{fig:store_rate}
\end{figure*}

\subsection{Transition Dissemination}
\label{sec:Transition Dissemination}

The attacker records interactions with $Victim$ in the form of transitions.
Each transition is represented as a tuple $(o_t, a_t, r_t, o_{t+1})$, where $o_t$ is the observation at time step $t$, $a_t$ is the joint action of $Adversary$, $r_t$ is the instantaneous reward, $o_{t+1}$ is the observation at the next time step, and $t \in [0, T-1]$, where $T$ is the time horizon.
Each agent in $Adversary$ maintains a set of replay buffers $\{ \mathcal{E}_k^i \}$ that stores the transitions for each subgame, where $i \in \mathcal{M}$ denotes the agent's identifier and $k \in \mathcal{K}$ denotes the subgame's identifier.

\noindent \textbf{Empirical Study.}
We introduce a concept of \emph{Occupancy Rate} (\emph{OR}), which quantifies the occurrence frequency of a subgame.
The occupancy rates are related to the number of transitions and serve as the foundation for constructing the transition dissemination table.
We conduct an empirical study to explore the relationship between occupancy rates and three partially observable limitations (please refer to \autoref{sec:Setup} for setup details).
The results in~\autoref{fig:store_rate} reveal two key observations.

\noindent \textbf{Observation 1.} (Heterogeneity Property)
\textit{The occupancy rate of subgames exhibit variations, which is further affected by the limitations of partial observability.}

Using the first column of~\autoref{fig:store_rate}(c) as an example, the occupancy rates for four subgames $\mathcal{G}{\alpha_0}, \mathcal{G}{\alpha_1}, \mathcal{G}{\alpha_2}$ and $\mathcal{G}{\alpha_3}$ are 0.04, 0.08, 0.24, and 0.65, respectively.
This highlights that some subgames, such as $\mathcal{G}{\alpha_2}$ and $\mathcal{G}{\alpha_3}$, occur frequently, leading to higher occupancy rates (0.24 and 0.65).
These subgames provide a sufficient number of transitions, contributing to a more comprehensive dataset for learning and analysis.
On the other hand, subgames like $\mathcal{G}{\alpha_0}$ and $\mathcal{G}{\alpha_1}$ occur infrequently, resulting in lower occupancy rates (0.04 and 0.08).
The infrequent occurrence of these subgames leads to a scarcity of transitions, which may present challenges for effective learning and decision-making within specific contexts.

\noindent \textbf{Observation 2.} (Dynamics Property)
\textit{Under distance limitations, the occupancy rate of subgames is influenced by variations in the attacker's policy.}

The occupancy rates remain stable under uncertainty and region limitations, but they exhibit significant shifts under distance limitations. For example, \emph{OR}$_0$ increases from 0.03 to 0.14, while \emph{OR}$_3$ decreases from 0.57 to 0.35.
The distance limitation is associated with an observable range represented by a circle centered around each agent in $Adversary$. Therefore, the observation by an agent is determined by its behavior pattern or policy.

The heterogeneity property underscores the non-uniform nature of occupancy rates across different subgames. We propose three methods aim to address this non-uniformity and determine occupancy rate values in such scenarios.

\noindent \textbf{Static Estimation.}
Under uncertainty limitations, if we assume that the uncertainty in the observation of any $Victim$'s agent is the same, $OR_k$ is considered as the probability of repeatedly observing \emph{N} agents in $Victim$ with exactly $k$ successes. The probability of successful observation is $\mu$, which can be introduced as priori knowledge or obtained from historical observation of $Victim$. Thus, the occupancy rate obeys a binomial distribution with parameters \emph{N} and $\mu$, \emph{i.e.},
\begin{equation}
OR_k = \tbinom{N}{k} \cdot \mu^k \cdot (1 - \mu)^{N - k},
\label{eq:set}
\end{equation}
where $\tbinom{N}{k}$ is the binomial coefficient, which represents the number of ways to choose $k$ successes from $N$ trials.

\noindent \textbf{Static Observation.}
Alternatively, the attacker calculates the occupancy rate for each subgame $\mathcal{G}_{\alpha_k}$ by counting the number of related transitions in all replay buffers.
\begin{equation}
OR_k = \frac{|\mathcal{E}_k|}{|\mathcal{E}_0| + |\mathcal{E}_1| + ... + |\mathcal{E}_{Sub - 1}|},
\label{eq:sot}
\end{equation}
where $|\mathcal{E}_{k}| = \sum_{i=1}^M |\mathcal{E}_{k}^i|$ and $|\mathcal{E}_{k}^i|$ indicates the number of transitions stored in $\mathcal{E}_{k}^i$. These transitions can either be direct interactions with $Victim$ or observations of $Victim$'s interactions with other MASs. Static observation does not rely on additional assumptions or prior knowledge and is applicable to all three types of limitations discussed in this paper.

\noindent \textbf{Dynamic Observation.}
The dynamics property reveals that occupancy rates may exhibit significant fluctuations and migration in a competitive environment. To account for this, the attacker can utilize \emph{Exponentially Weighted Averages} to accommodate the dynamics of occupancy rates:
\begin{equation}
OR_{k_t} = \beta \cdot OR_{k_{t-1}} + (1 - \beta) \cdot OR'_{k_t},
\label{eq:dt}
\end{equation}
where $OR_{k_{t-1}}$ denotes the current weighted average of $\mathcal{G}_{\alpha_k}$'s occupancy rate, $\beta$ denotes the weight of historical transition (\emph{e.g.}, $\beta$ = 0.9 means that the occupancy rate is approximately to the average of 10 episodes), and $OR'_{k_t}$ denotes the occupancy rate obtained from the current episode statistics. Dynamic observation is applicable to all the three limitations and does not require additional assumptions or prior knowledge.
\autoref{sec:Attack_Performance} elucidates the strengths and weaknesses of these three methods through experiments and analysis, guiding on their selection.

\begin{figure}
\centering
\includegraphics[height=2.5 in]{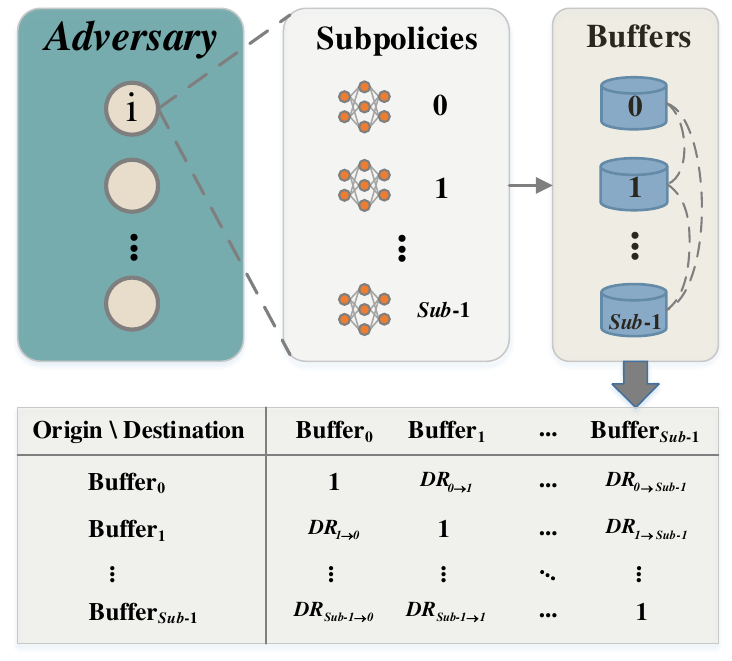}
\caption{Each agent in $Adversary$ maintains a transition dissemination table.}
\label{fig:table}
\end{figure}

\noindent \textbf{Transition Dissemination Table.} The attacker determines the \emph{Dissemination Rate} (\emph{DR}) between each pair of replay buffers based on the occupancy rate. As shown in~\autoref{fig:table}, for each agent $i \in \mathcal{M}$, we define $DR_{\hat{k} \rightarrow k} \in [0, 1]$ as the probability of transition transmission from buffer $\mathcal{E}^i_{\hat{k}}$ to buffer $\mathcal{E}_k^i$.

\emph{DR}s are determined by four factors:
(1) $DR_{\hat{k} \rightarrow k}$ is negatively correlated with the occupancy rate of the destination buffer $\mathcal{E}_k^i$ since the buffer with sufficient transitions does not require additional transitions.
(2) $DR_{\hat{k} \rightarrow k}$ is negatively correlated with the distance between $\mathcal{G}_{\alpha_{\hat{k}}}$ and $\mathcal{G}_{\alpha_k}$ since the transitions with higher similarity are provided between buffers close to each other.
(3) $DR_{\hat{k} \rightarrow k}$ is positively correlated with the number of constructed subgames since the mean value of occupancy rates decreases as \emph{Sub} increases.
(4) $DR_{\hat{k} \rightarrow k}$ is positively correlated with the dispersion of occupancy rates since high dispersion is prone to multiple occupancy rates with lower values. 
In summary, we define $DR_{\hat{k} \rightarrow k}$ as
\begin{equation}
DR_{\hat{k} \rightarrow k} = \\
\begin{cases}
\mathrm{clip}((\lambda - OR_k + \sigma)^{\frac{|\hat{k}-k|}{\sqrt{Sub}}}, 0, 1), & \mathrm{if} \ OR_k \leq \lambda\\
\sigma^{\frac{|\hat{k}-k|}{\sqrt{Sub}}}, & \mathrm{if} \ OR_k > \lambda
\end{cases}
\label{eq:edr}
\end{equation}
where $\lambda$ is an adjustment parameter positively correlated with the complexity of the two-team competitive environment. $\sigma$ is the standard deviation of all occupancy rates and is used to measure their dispersion. $|\hat{k}-k|$ indicates the distance between two subgames. $\sqrt{Sub}$ indicates that as the number of constructed subgames increases, the transition dissemination between buffers will become more frequent, meaning that the value of $DR_{\hat{k} \rightarrow k}$ will increase. We use the clip function to limit the value of $DR_{\hat{k} \rightarrow k}$ to the range [0,1].

All \emph{DR}s collectively form a transition dissemination table.
When a new transition is generated, the agent allocates it to replay buffers based on the probabilities recorded in this table.
If static estimation or static observation is employed to determine occupancy rates, this table is static. In contrast, if dynamic observation is used to determine occupancy rates, this table changes dynamically.
The impact of transition dissemination is presented in~\autoref{fig:heatmap} in the appendix.

\subsection{Subpolicy Training}
\label{sec:Subpolicy Training}

Each agent $i \in \mathcal{M}$ maintains a set $\{ \pi_{\alpha_k}^i \}_{k \in \mathcal{K}}$ consisting of all subpolicies. We perform \emph{Policy Meritocracy} (\emph{PM}) to preserve top-performing subpolicies based on the harmonic mean of their test performance across $L$ metrics, mitigating performance fluctuations caused by non-stationarity.
\begin{equation}
PM_k^i = \frac{L}{\sum\limits_{l=1}^L 1 / \eta_{k_l}^i},
\label{eq:pm}
\end{equation}
where $\eta_{k_l}^i$ indicates the test performance of the subpolicy $\pi_{\alpha_k}^i$ with respect to metric $l$. In the policy pool, only one subpolicy is retained for each subgame to minimize storage overhead.
Replacements occur when a subpolicy with superior test performance emerges. The detailed update process for the subpolicies is outlined in~\autoref{alg:Subpolicy_Training} in the appendix.

\begin{figure*}[t]
\centering
\includegraphics[height=2.8 in]{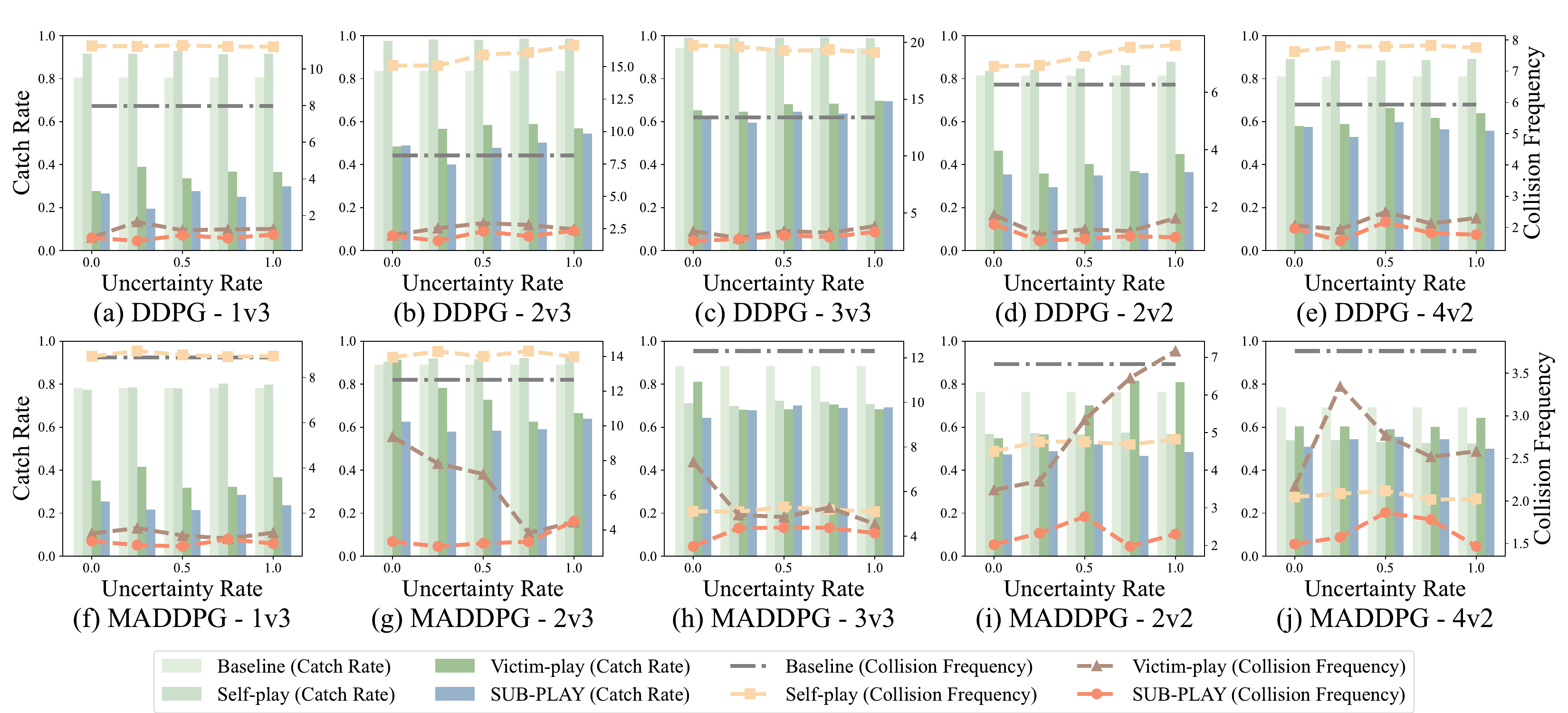}
\caption{The attack performance of \emph{SUB-PLAY} under uncertainty limitations in Predator-prey.}
\label{fig:uncertainty}
\end{figure*}

\subsection{Policy Combination}
\label{sec:Policy Combination}

The attacker combines all subpolicies $\{ \pi_{\alpha_k}^i \}_{i \in \mathcal{M}, k \in \mathcal{K}}$ to generate the final adversarial policy. Specifically, at each time step of the deployment phase, $Adversary$ obtains a joint observation $(o_1, o_2, ... , o_M)$ and then outputs a joint action $(a_1, a_2, ... , a_M)$.
The subgame construction ensures that each observation exclusively belongs to a single subgame, without any overlap or intersection between the subgames, \emph{i.e.}, $\Omega_{\hat{k}}^i \cap \Omega_k^i = \varnothing$, where $\hat{k} \neq k$ ($\hat{k}, k \in \mathcal{K}$), $i \in \mathcal{M}$.
Furthermore, unlike backdoor attacks~\cite{goldblum2022dataset, wang2021backdoorl}, the attacker can modify the program structure without concerns about stealthiness. Hence, we propose that the attacker can implement the policy combination in a hard-coded manner.

Taking the distributed MARL as an example, each agent $i \in \mathcal{M}$ determines which subgame a received observation $o_i$ belongs to and subsequently selects the corresponding subpolicy to guide its action decision-making:
\begin{equation}
\pi_\alpha^i = \\
\begin{cases}
\pi_{\alpha_0}^i, & \mathrm{if} \ o_i \in \Omega_0^i\\
...\\
\pi_{\alpha_{Sub-1}}^i, & \mathrm{if} \ o_i \in \Omega_{Sub-1}^i
\end{cases}
\label{eq:subpolicy}
\end{equation}

This approach has two advantages: it provides a straightforward logic and greater flexibility, allowing each subgame to employ a separate MARL algorithm.
The implementation of policy combination can be found in~\autoref{alg:Policy Combination} in the appendix.

\section{Evaluation}
\label{sec:Evaluation}

This section first introduces the evaluation setup and then evaluates \emph{SUB-PLAY} through six perspectives: attack performance, ablation study, transferability, scalability, overhead, and potential defenses.

\subsection{Setup}
\label{sec:Setup}

\noindent \textbf{Environment.}
We adopt two-dimensional environments, Predator-prey and World Communication, within the Multi Particle Environments (MPE)~\cite{lowe2017multi} framework developed by OpenAI.
MPE is a gym-based benchmark designed for cooperative, competitive, and mixed MARL tasks.

\begin{itemize}
    \item Predator-prey. There are \emph{N} slower predators controlled by $Victim$ and \emph{M} preys controlled by $Adversary$. The predators cooperate to collide with preys, while the preys cooperate to avoid collisions. The environment is initialized randomly at the beginning of each episode, including the positions of agents and obstacles.
    \item World Communication. This environment shares similarities with the Predator-prey setup but includes additional features. (1) There are \emph{M} foods that preys are rewarded for being close to. (2) The environment randomly initializes a forest, making all agents invisible within it initially. (3) A leader predator exists, having full visibility of all agents and the ability to communicate with other predators to enhance cooperation.
\end{itemize}

The state and action spaces of both environments are continuous.
The evaluation of the attack performance is conducted in five different scenarios (\emph{M}v\emph{N}): 1v3, 2v3, 3v3, 2v2, and 4v2.

\noindent \textbf{Partial Observability Implementation.}
Observation is a multi-dimensional vector. We additionally introduce a Mask vector consisting of 0s and 1s, which has the same dimensions as Observation. The Mask is determined by specific rules for partial observability. We multiply each element of Observation with the corresponding element of the Mask, and the result is partial observation.
For specific implementation, please consult \autoref{alg:Partially Observable Implementation} in the appendix.

In Predator-prey, we conduct evaluations under both uncertainty and distance limitations.
The uncertainty rate (\emph{i.e.} $1 - \mu$ in~\autoref{eq:set}) ranges from $\{0.00, 0.25, 0.50, 0.75, 1.00 \}$.
For example, $1 - \mu = 0.00$ indicates complete observability, and the mask array consists of all 1s.
The observable distance ranges from $\{0.5, 1.0, 1.5, 2.0 \}$.

In World Communication, we conduct evaluations under region limitations.
This environment has implemented partial observability, where agents located within a specific region are not visible (a forest with a fixed size but randomly initialized positions).

\noindent \textbf{Implementation Details.}
Our evaluations are conducted on four servers with Intel(R) Xeon(R) CPU E5-2650 v4 @ 2.20GHz, 32GB RAM.
Python and PyTorch are used for code implementation.

Selected MARL algorithms include DDPG and MADDPG, serving as representatives for DTDE and CTDE architectures, respectively.
DDPG and MADDPG are actor-critic algorithms commonly used in reinforcement learning. These algorithms consist of an actor network and a critic network.
We utilizes a two-layer ReLU MLP with 128 units in each layer to parameterize all policies. The actor network's output layer incorporates a Tanh activation function.
For weight initialization, we use Xavier normal with a gain of 1.0 for all layers in both victim and adversarial policy training.
Biases are initialized with zeros.
The chosen optimizer is Adam, with a learning rate of 0.001 and $\epsilon$ set to $10^{-8}$.

To ensure smooth policy updates, we employ \emph{Exponential Moving Average} (\emph{EMA}) with a decay rate of 0.95.
Random noise sampled from a normal distribution with a standard deviation of 0.01 is added to the output actions to promote exploration.
The discount factor $\gamma$ for RL is set to 0.95.
The adjustment parameter $\lambda$ in~\autoref{eq:edr} is set to 0.5. All reported results are averaged over 1,000 test runs.
For additional implementation details, parameter settings, and training results of $Victim$s, please refer to \autoref{app:Additional Evaluation Details}.

\begin{figure*}
\centering
\includegraphics[height=2.8 in]{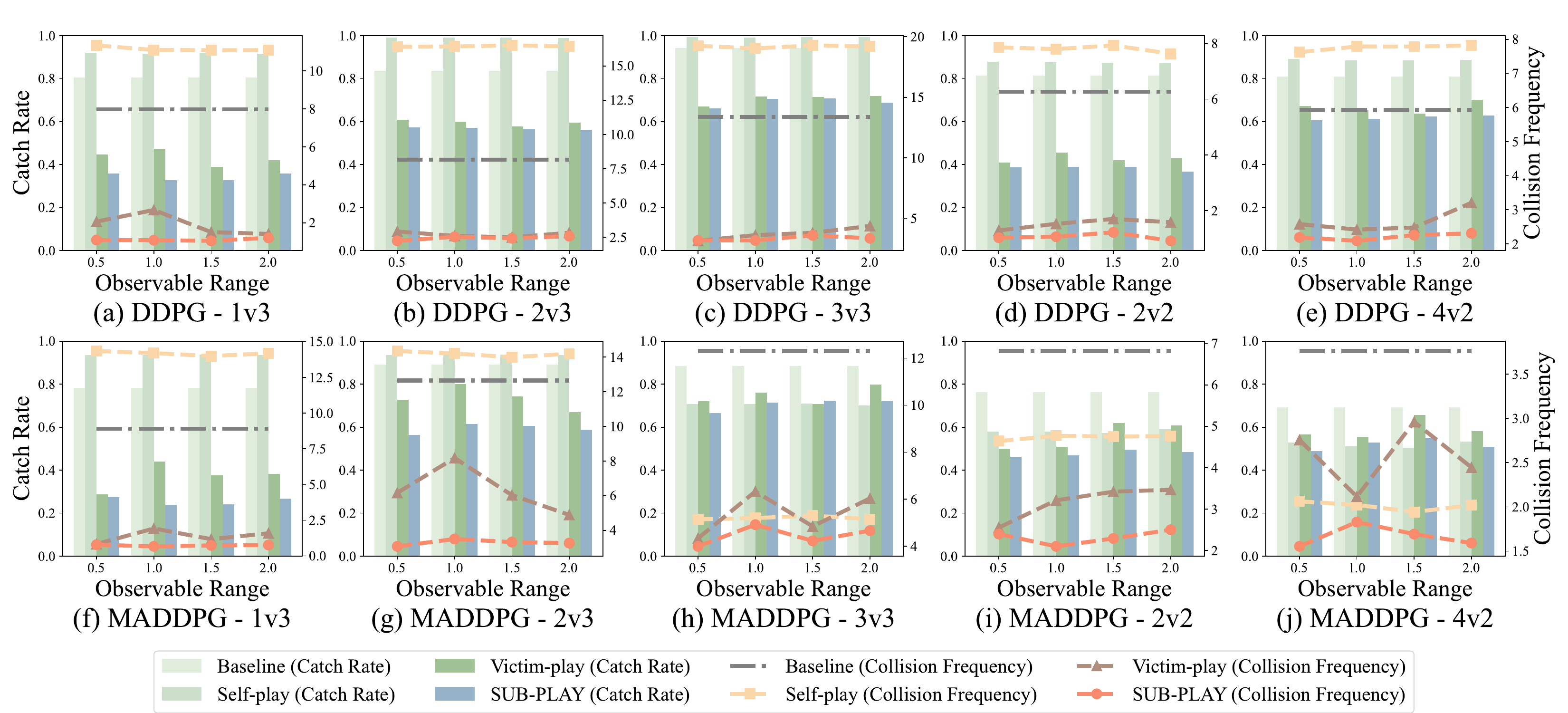}
\caption{The attack performance of \emph{SUB-PLAY} under distance limitations in Predator-prey.}
\label{fig:distance}
\end{figure*}

\begin{figure}
\centering
\includegraphics[height=1.55 in]{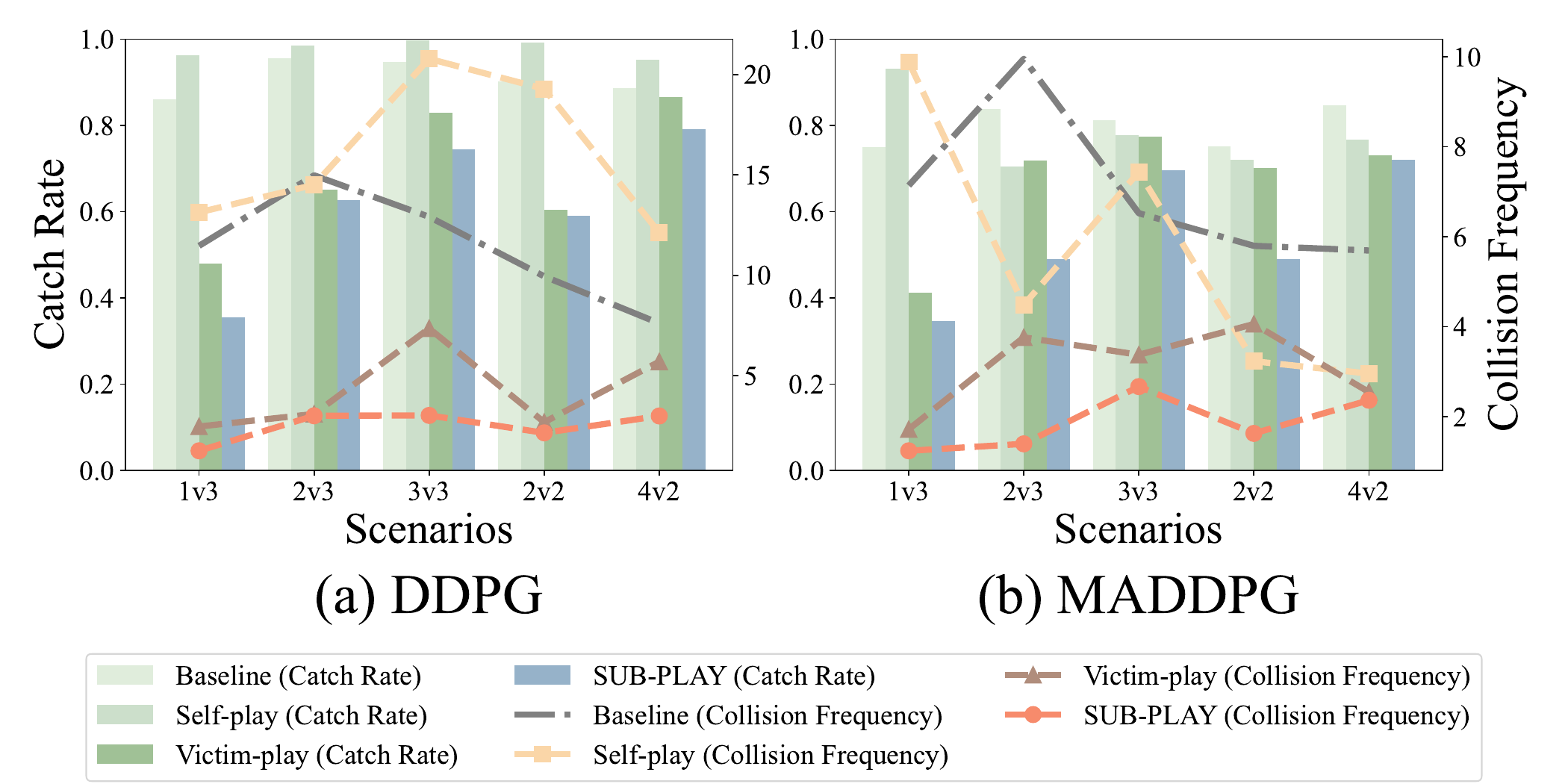}
\caption{The attack performance of \emph{SUB-PLAY} under region limitations in World Communication.}
\label{fig:region}
\end{figure}

\noindent \textbf{Comparison Methods.}
(1) Baseline: $Victim$'s normal performance on a specific task. Specifically, $Adversary$ deploys a heuristic policy where the agent's movement is characterized by a fixed speed and direction, with random updates after collisions.
(2) \emph{Self-play}~\cite{bansal2018emergent, berner2019dota}: Similar to the process of victim training, this setup grants the attacker complete access to the environment.
The attacker randomly initializes both $Adversary$ and $Victim$, allowing them to compete with each other and undergo updates. Ultimately, $Adversary$ is retained to carry out the attack.
(3) \emph{Victim-play}~\cite{gleave2020adversarial, wu2021adversarial, guo2021adversarial, wang2023adversarial, liu2023rethinking}: Apart from substituting the algorithm with MARL, retain the other fundamental settings of this framework (refer to~\autoref{sec:Adversarial_Policy}), specifically fixing $Victim$ and updating $Adversary$.

\noindent \textbf{Metrics.}
We adopt \emph{Catch Rate} (\emph{CR}) and \emph{Collision Frequency} (\emph{CF}) as evaluation metrics.
\begin{equation}
CR = \frac{Num_c}{Num_e},
\end{equation}
where $\emph{Num}_\emph{c}$ indicates the number of episodes where the $Adversary$ is caught; $\emph{Num}_\emph{e}$ indicates the total number of episodes.
\begin{equation}
CF = \frac{1}{Num_e} \sum_{e = 1}^{Num_e} \sum_{j = 1}^N Num_{ej},
\end{equation}
where $\emph{Num}_\emph{ej}$ denotes the number of collisions between the \emph{j}th predator and the prey in a specific episode \emph{e}.

The attacker aims to evade pursuit and minimize $Victim$'s \emph{CR} and \emph{CF}. Policy meritocracy is based on the harmonic mean of these two metrics,
\begin{equation}
PM = 2 \cdot \frac{CR \cdot CF}{CR + CF}.
\label{eq:pm1}
\end{equation}

We calculate the average improvement in attack performance of \emph{SUB-PLAY} compared to \emph{Victim-play} by
\begin{equation}
\frac{(PM_B - PM_S) - (PM_B - PM_V)}{PM_B - PM_V},
\label{eq:delta}
\end{equation}
where $PM_B$, $PM_V$, and $PM_S$ represent the victim's average performance when the attacker executes Baseline, \emph{Victim-play}, and \emph{SUB-PLAY}, respectively.

\subsection{Attack Performance}
\label{sec:Attack_Performance}

We evaluate the attack performance of \emph{SUB-PLAY} across different environments and limitations. The training process of \emph{SUB-PLAY} can be found in~\autoref{fig:curve_attack} in the appendix.

\noindent \textbf{Uncertainty Limitation.}
Under uncertainty limitations, \emph{SUB-PLAY}, on average, reduces the victim's performance to 51.98\% of the baseline (see~\autoref{fig:uncertainty}).
Moreover, \emph{SUB-PLAY} outperforms the other two methods and minimizes the victim's catch and collision rates in 94.0\% (47/50) and 98.0\% (49/50) scenarios.
Compared to \emph{Victim-play}, \emph{SUB-PLAY} demonstrates an average improvement of 32.22\% in attack performance (refer to~\autoref{eq:delta}).

The results also show that when the MARL algorithm is MADDPG (resulting in increased input dimensionality due to information sharing among agents), \emph{SUB-PLAY} demonstrates more stable attack performance, showcasing its potential to handle more complex environments.
In addition, an unexpected outcome is that the attack remains effective even when the uncertainty is set to 1.00 (\emph{i.e.}, the attacker has no observations of the victim).
According to Silver \emph{et al.}~\cite{silver2021reward}, maximizing rewards is sufficient to drive intelligent behavior.
In our scenarios, this suggests that the attacker can find adversarial policies by focusing on maximizing the rewards obtained, even without direct observation of the victim.

\noindent \textbf{Distance Limitation.}
Under distance limitations, \emph{SUB-PLAY}, on average, reduces the victim's performance to 55.71\% of the baseline (see~\autoref{fig:distance}).
Moreover, \emph{SUB-PLAY} outperforms the other two methods and minimizes the victim's catch and collision rates in 97.5\% (39/40) scenarios.
Compared to other methods, the impact of the observation range on \emph{SUB-PLAY} is relatively minor, indicating that \emph{SUB-PLAY} demonstrates better adaptability to dynamic environments.
Compared to \emph{Victim-play}, \emph{SUB-PLAY} demonstrates an average improvement of 27.16\% in attack performance.

\noindent \textbf{Region Limitation.}
Under region limitations, \emph{SUB-PLAY} reduces the victim's performance to an average of 59.07\% of the baseline (see~\autoref{fig:region}).
Moreover, \emph{SUB-PLAY} outperforms the other two methods and minimizes the victim's catch and collision rates in 100.0\% (10/10) scenarios.
Compared to \emph{Victim-play}, \emph{SUB-PLAY} demonstrates an average improvement of 50.22\% in attack performance.
These results indicate that \emph{SUB-PLAY} exhibits more significant attack potential in complex environments.

Furthermore, we compare \emph{SUB-PLAY} with two additional variants of \emph{Victim-play}~\cite{guo2021adversarial, wang2023adversarial}, further validating \emph{SUB-PLAY}'s attack performance in partially observable environments.
Details regarding the settings of these two variants and the final results are provided in the~\autoref{app:Additional Comparison}.

\noindent \textbf{Visualization Results.}
The visualization results (\autoref{fig:visualization} in the appendix) show that in Predator-prey, the preys tend to flee to the edge of the map at the maximum speed from different directions and then stay while trying to bypass predators and obstacles. They quickly get rid of collisions if they occur.

\begin{figure}
\centering
\includegraphics[height=1.9 in]{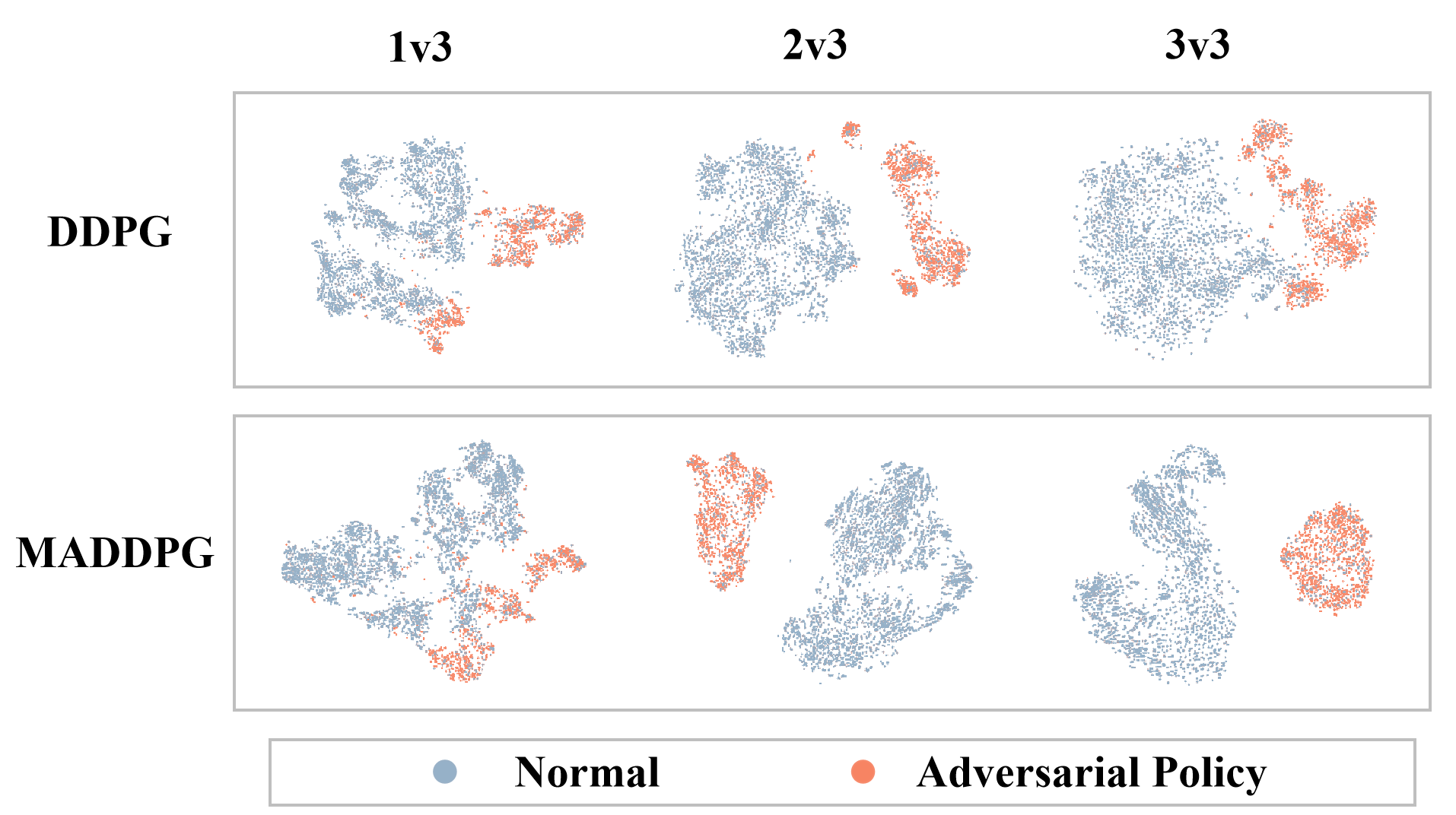}
\caption{t-SNE activations of the victim when playing against different opponents.}
\label{fig:tsne1}
\end{figure}

Similarly, in World Communication, the preys do not hide in the forest or approach foods for additional rewards. These observations demonstrate that adversarial policies effectively utilize the speed advantage of the preys to evade predators, mitigating the disadvantages of partial observability. t-SNE (see~\autoref{fig:tsne1}) shows that the activations of the victim's policy network are significantly different when facing a normal opponent compared to an adversarial policy. More results can be found in~\autoref{fig:tsne2} in the appendix.

\begin{table*}
\scriptsize
\renewcommand\arraystretch{1.2}
\caption{The impact of three occupancy rate determination methods on attack performance under three partial observability limitations. The performance is measured by two metrics (CR$\downarrow$/CF$\downarrow$). Acronyms: Static Estimation (SE), Static Observation (SO), Dynamic Observation (DO).}
\label{tab:occupancy}
\centering
\begin{tabular}{|c|c|ccc|ccc|} \hline
\multicolumn{2}{|c|}{\multirow{2}*{\textbf{Limitations}}} & \multicolumn{3}{c|}{\textbf{DDPG}} & \multicolumn{3}{c|}{\textbf{MADDPG}} \\ \cline{3-8}
\multicolumn{2}{|c|}{\multirow{2}{*}{}} & \textbf{SE} & \textbf{SO} & \textbf{DO} & \textbf{SE} & \textbf{SO} & \textbf{DO}\\ \hline
\multirow{5}*{\textbf{Uncertainty}} & 0.00 & \textbf{0.489 / 1.969} & 0.527 / 2.115 & 0.571 / 2.527 & 0.626 / \textbf{3.327} & 0.622 / 4.007 & \textbf{0.606} / 3.846 \\
\multirow{5}{*}{} & 0.25 & \textbf{0.401 / 1.573} & 0.538 / 2.205 & 0.575 / 2.532 & \textbf{0.579 / 3.053} & 0.607 / 3.506 & 0.725 / 6.460 \\
\multirow{5}{*}{} & 0.50 & \textbf{0.477 / 2.309} & 0.530 / 2.430 & 0.619 / 3.362 & \textbf{0.583 / 3.228} & 0.605 / 3.389 & 0.697 / 6.209 \\
\multirow{5}{*}{} & 0.75 & \textbf{0.502 / 1.927} & 0.532 / 2.395 & 0.580 / 2.475 & \textbf{0.591 / 3.318} & 0.614 / 3.909 & 0.639 / 3.904 \\
\multirow{5}{*}{} & 1.00 & \textbf{0.543 / 2.345} & 0.566 / 2.563 & 0.592 / 2.919 & \textbf{0.638} / 4.506 & 0.647 / \textbf{4.019} & 0.687 / 4.880 \\ \hline
\multirow{4}{*}{\textbf{Distance}} & 0.5 & - & 0.598 / 2.554 & \textbf{0.574 / 2.240} & - & 0.609 / 3.335 & \textbf{0.563 / 3.075} \\
\multirow{4}{*}{} & 1.0 & - & 0.657 / 3.059 & \textbf{0.571 / 2.529} & - & 0.647 / 4.148 & \textbf{0.614 / 3.500} \\
\multirow{4}{*}{} & 1.5 & - & 0.609 / 3.025 & \textbf{0.564 / 2.421} & - & 0.667 / 4.430 & \textbf{0.606 / 3.324} \\
\multirow{4}{*}{} & 2.0 & - & 0.620 / 2.786 & \textbf{0.561 / 2.579} & - & 0.654 / 4.464 & \textbf{0.589 / 3.264} \\ \hline
\textbf{Region} & 1 & - & 0.664 / \textbf{2.917} & \textbf{0.626} / 2.998 & - & 0.594 / 1.855 & \textbf{0.489 / 1.397} \\
\hline
\end{tabular}
\end{table*}

\noindent \textbf{Proactive Masking.} Intuitively, partial observability poses a challenge for adversarial policy generation.
However, evaluations indicate that proactive masking of environment observations may enhance the attack performance in some scenarios.
For example, in \autoref{fig:uncertainty}(a), the attack performance of \emph{SUB-PLAY} is superior when the uncertainty rate is 0.25 compared to the fully observable scenario.
The reason is that the minor partial observability corresponds to simplifying the input to the attacker.
This suggests that proactive masking may facilitate early training of adversarial policies in complex environments.
Recent work has introduced curriculum learning to address sparse rewards in adversarial policy generation \cite{wang2023adversarial}, but it requires the access to different versions of victims.
In contrast, proactive masking does not rely on a similar assumption.

\noindent \textbf{Impact of Occupancy Rates.}
To compare the impact of the three methods proposed in \autoref{sec:Transition Dissemination} for determining occupancy rates, we evaluate them under three partial observability limitations (static estimation, static observation, and dynamic observation).
In the static observation, the attacker pre-observes the victim for 1000 episodes, while in the dynamic observation, the parameter $\beta$ is set to 0.9.

\autoref{tab:occupancy} shows that \emph{SUB-PLAY} achieves optimal attack performance with static estimation under uncertainty limitations. This is because static estimation provides more accurate occupancy rate estimations.
Furthermore, \emph{SUB-PLAY} performs superior attacks using dynamic observation under distance and region limitations.
The former is due to the change in occupancy rates as the adversarial policy updates, while the latter is attributed to the additional dynamics introduced by the randomly initialized position of the forest in World Communication.

Therefore, in all evaluations, we set that the attacker adopts static evaluation under uncertainty limitations and dynamic observation under distance and region limitations.

\begin{table*}
\scriptsize
\renewcommand\arraystretch{1.2}
\caption{The ablation results of components in \emph{SUB-PLAY} measured by two metrics (CR$\downarrow$/CF$\downarrow$). Acronyms: Subgame Construction (SC), Transition Dissemination (TD), Policy Meritocracy (PM).}
\label{tab:ablation}
\centering
\begin{tabular}{|c|ccccc|} \hline
\multirow{2}{*}{\textbf{Methods}} & \multicolumn{5}{c|}{\textbf{Limitations}}\\ \cline{2-6}
\multirow{2}{*}{} & Uncertainty (0.25) & Uncertainty (0.50) & Distance (0.5) & Distance (2.0) & Region (1)\\ \hline
\emph{Self-play} & 0.920 / 14.280 & 0.916 / 13.998 & 0.936 / 14.349 & 0.935 / 14.187 & 0.704 / 4.486\\
\emph{Victim-play} & 0.782 / 7.823 & 0.727 / 7.215 & 0.728 / 6.163 & 0.670 / 4.891 & 0.718 / 3.763\\
\emph{SUB-PLAY} (SC) & 0.830 / 8.402 & 0.759 / 7.604 &  0.765 / 6.296 &  0.708 / 5.982  &  0.835 / 6.563\\
\emph{SUB-PLAY} (SC+TD) & 0.617 / 3.740 & 0.627 / 4.438 &  0.700 / 6.552 & 0.672 / 4.675 &  0.688 / 3.309\\
\emph{SUB-PLAY} (SC+PM) & 0.731 / 6.059 & 0.708 / 6.318 &  0.735 / 6.113 & 0.677 / 4.576 &  0.561 / 1.634\\
\emph{SUB-PLAY} (SC+TD+PM) & \textbf{0.579 / 3.053} & \textbf{0.583 / 3.228} & \textbf{0.563 / 3.075} & \textbf{0.589 / 3.264} & \textbf{0.489 / 1.397}\\
\hline
\end{tabular}
\end{table*}

\subsection{Ablation Study}

Unless additional specifications exist, the subsequent evaluations use the 2v3 scenario, and the MARL algorithm is set to MADDPG.

\noindent \textbf{Component Evaluation.}
We perform an ablation study to assess the contribution of each component (subgame construction, transition dissemination, and policy meritocracy in subpolicy training) in \emph{SUB-PLAY}. 
\autoref{tab:ablation} demonstrates that when the subgame construction is applied in isolation, the attack performance is inferior. 
However, when transition dissemination is also performed, the attack performance is significantly improved.
This highlights the crucial role of transition dissemination in enhancing performance in partially observable environments.
Policy meritocracy also contributes to improved attack performance and is compatible with subgame construction and transition dissemination.

\begin{figure}
\centering
\includegraphics[height=1.7 in]{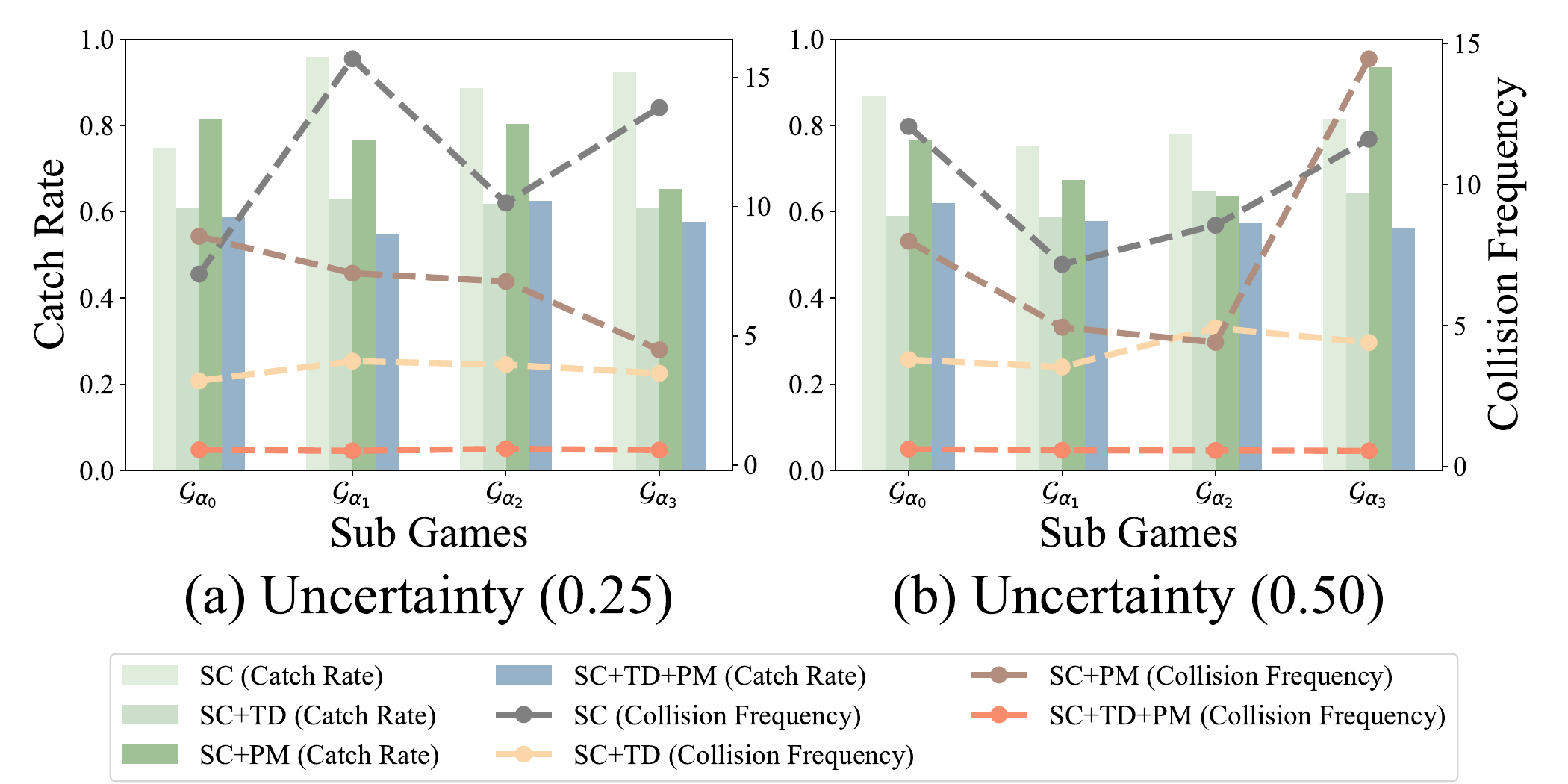}
\caption{The attack performance of subpolicies under uncertainty limitations.}
\label{fig:Sub}
\end{figure}

\noindent \textbf{Subgame Evaluation.}
We continue to explore the reasons behind the performance improvements achieved by \emph{SUB-PLAY}.
\autoref{fig:Sub} presents the attack performance of subpolicies in their respective corresponding subgames under uncertainty limitations. The results reveal that when the attacker only applies subgame construction and policy meritocracy, the resulting adversarial policies exhibit substantial performance differences across different subgames. However, with the inclusion of transition dissemination, all subpolicies show notable improvements in their attack performance. Furthermore, the performance disparities among the subpolicies are significantly reduced.

\subsection{Transferability}

In real-world scenarios, the limitations on partial observability are not static; for instance, environmental factors like weather can impact sensor performance, leading to variations in the observable distance for the adversary agents.
Therefore, we evaluate the transferability of adversarial policies across different parameter settings within the same limitation.

\begin{figure} [tb]
\centering
\includegraphics[height=3 in]{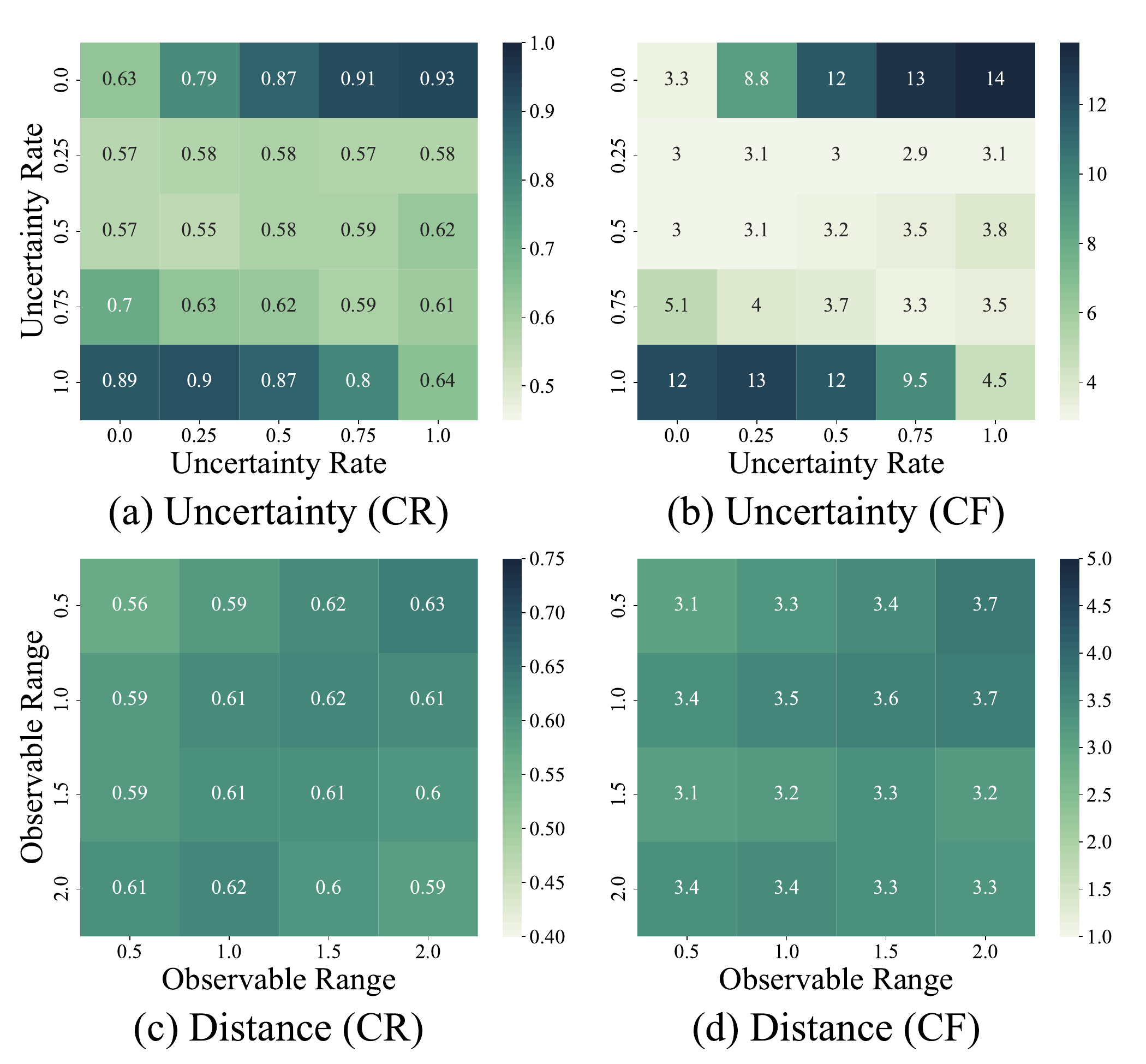}
\caption{The transferability of adversarial policies across different scenarios. The vertical coordinate represents the limitation set in the training phase and the horizontal coordinate represents the limitation set in the testing phase.}
\label{fig:Transferability}
\end{figure}

\autoref{fig:Transferability}(a)-(b) illustrate that adversarial policies can be transferred in specific scenarios (\emph{i.e.}, when the uncertainty rate is 0.25, 0.50, and 0.75).
However, transferability diminishes when the uncertainty rate is 0.00 or 1.00.
This is due to the significant differences in the modeling between fully unobservable and fully observable scenarios compared to partially observable scenarios.
Additionally, subgame construction and transition dissemination are ineffective in these two cases.
\autoref{fig:Transferability}(c)-(d) demonstrate that adversarial policies exhibit transferability under distance limitations.
In conclusion, adversarial policies derived from \emph{SUB-PLAY} demonstrate transferability in similar partially observable environments.

However, the transferability of adversarial policies encounters limitations when applied to victims with substantial differences.
\autoref{tab:random} illustrates that the adversarial policy proves ineffective in reducing the catch rate of heuristic victims, although it succeeds in diminishing their collision frequency.
The reduction in collision frequency is attributed to the adversarial policy's capacity to swiftly escape captures, whereas heuristic attackers frequently engage in repeated collisions.
The inability to decrease the victims' catch rate results from adversaries consistently moving at the maximum speed in the adversarial policy, thereby increasing the probability of collision -- especially when preys are slow, potentially concluding the episode before colliding with predators. Moreover, preys lose the ability to evade nearby predators.

\subsection{Scalability}
\label{sec:Scalability}

\emph{SUB-PLAY} is scalable, meaning the attacker can adjust the granularity of subgame construction (\autoref{sec:Subgame Construction}) based on the scenario and requirements.
\autoref{eq:sub} demonstrates the most potent attack by constructing \emph{Sub = N+1} sub-games against \emph{N} victim agents. However, in resource-constrained environments, the attacker can adopt a coarser granularity, i.e., \emph{Sub $<$ N+1}.

\begin{figure}
\centering
\includegraphics[height=1 in]{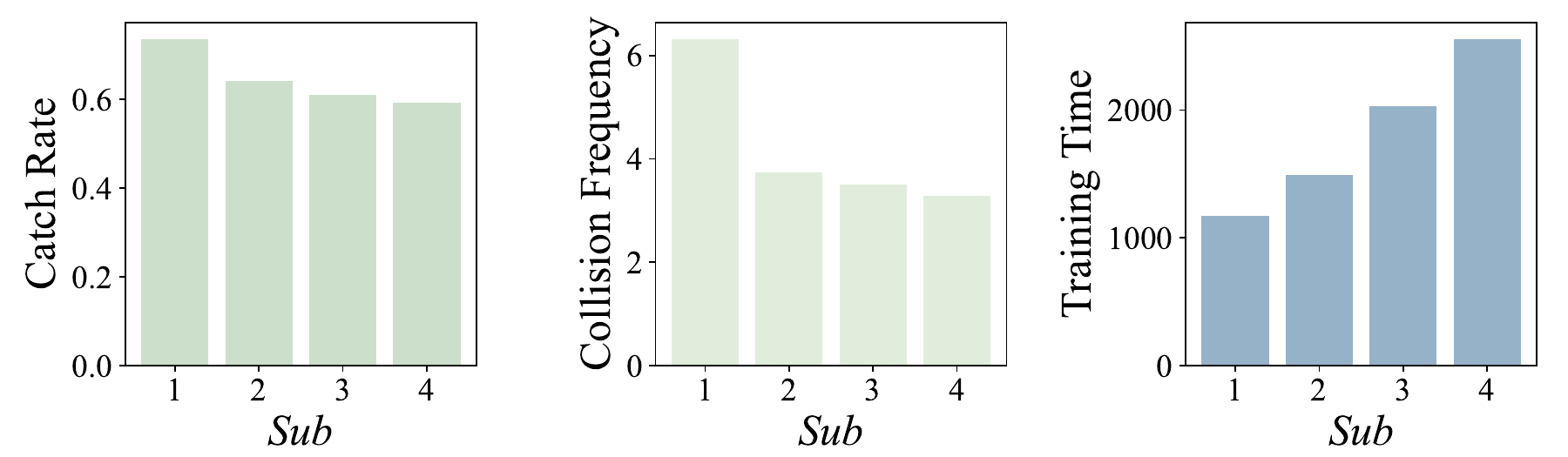}
\caption{Scalability evaluation.}
\label{fig:scalability}
\end{figure}

We validate the scalability of \emph{SUB-PLAY} in Predator-prey (algorithm = MADDPG, distance = {0.5, 1.0, 1.5, 2.0}, scenarios = 2v3), constructing 1, 2, 3, and 4 subgames. \autoref{fig:scalability} illustrates that under four different granularity settings, \emph{SUB-PLAY} can decrease the victim's catch rate to 0.736, 0.641, 0.611, and 0.593, and the collision frequency decreases to 6.315, 3.738, 3.504, and 3.291, respectively.
The results indicate that finer granularity leads to better attack effectiveness, while the rate of performance improvement diminishes as the number of subgames increases, gradually approaching saturation.
Additionally, the training time of \emph{SUB-PLAY} shows a positive correlation with the number of subgames (1171.97s, 1495.39s, 2031.03s, and 2556.17s, respectively), indicating that attackers can manipulate the granularity of subgame construction to scale \emph{SUB-PLAY} to more complex environments.

\subsection{Overhead}

The main overhead of adversarial policy-based attacks arises from three aspects: interaction costs, training costs, and decision delays.
\emph{SUB-PLAY} avoids extra interaction costs (\autoref{sec:Transition Dissemination}) and decision delays (\autoref{sec:Policy Combination}) but incurs additional training costs due to subgame construction (\autoref{sec:Subgame Construction}) and separate training for each subgame (\autoref{sec:Subpolicy Training}), which scales linearly with the number of subgames.

We explore the training time of \emph{SUB-PLAY} in Predator-prey (algorithm = {DDPG, MADDPG}, distance = 0.5, scenarios = {1v3, 2v3, 3v3, 2v2, 4v2}).
The results in \autoref{tab:training_time} indicate that under the condition of maximizing the number of subgames (\emph{Sub = N+1}), the average training times of \emph{SUB-PLAY} in distributed and centralized MARL algorithms are 1529s and 2781s, respectively (1.73 and 2.02 times that of \emph{Victim-play}).
Nevertheless, the training time of \emph{SUB-PLAY} remains significantly lower than that of the well-trained victim (only 2.39\%). Additionally, as demonstrated in \autoref{sec:Scalability}, the attacker can adjust the granularity of subgame construction to reduce training costs.

\begin{table}
\scriptsize
\renewcommand\arraystretch{1.2}
\caption{The attack performance of adversarial policies with heuristic victims. The performance is measured by two metrics (CR$\downarrow$/CF$\downarrow$).}
\label{tab:random}
\centering
\begin{tabular}{|c|c|c|c|} \hline
\multicolumn{2}{|c|}{\textbf{Attaker}}  & Heuristic & \emph{SUB-PLAY} \\ \hline
\multicolumn{2}{|c|}{\textbf{Victim}} & \multicolumn{2}{c|}{Heuristic} \\ \hline
\multirow{2}*{\textbf{Uncertainty}} & 0.25 & 0.261 / 0.699 & 0.307 / 0.599 \\
\multirow{2}{*}{} & 0.50 & 0.259 / 0.694 & 0.286 / 0.515 \\ \hline
\multirow{2}{*}{\textbf{Distance}} & 0.5 & 0.225 / 0.613 & 0.275 / 0.508 \\
\multirow{2}{*}{} & 2.0 & 0.265 / 0.692 & 0.313 / 0.600 \\ \hline
\textbf{Region} & 1 & 0.304 / 0.802 & 0.439 / 0.807 \\
\hline
\multicolumn{2}{|c|}{\textbf{Victim}} & \multicolumn{2}{c|}{MADDPG} \\ \hline
\multirow{2}*{\textbf{Uncertainty}} & 0.25 & 0.900 / 12.643 & 0.579 / 3.053 \\
\multirow{2}{*}{} & 0.50 & 0.918 / 13.258 & 0.583 / 3.228 \\ \hline
\multirow{2}{*}{\textbf{Distance}} & 0.5 & 0.942 / 13.540 & 0.563 / 3.075 \\
\multirow{2}{*}{} & 2.0 & 0.920 / 13.224 & 0.589 / 3.264 \\ \hline
\textbf{Region} & 1 & 0.837 / 9.954 & 0.489 / 1.397 \\
\hline
\end{tabular}
\end{table}

\subsection{Potential Defenses}
\label{sec:defenses}

Adversarial policies pose significant challenges to the real-world deployment of MARL. In response, we explore viable defense approaches to mitigate these security threats.

\begin{figure}
\centering
\includegraphics[height=1.15 in]{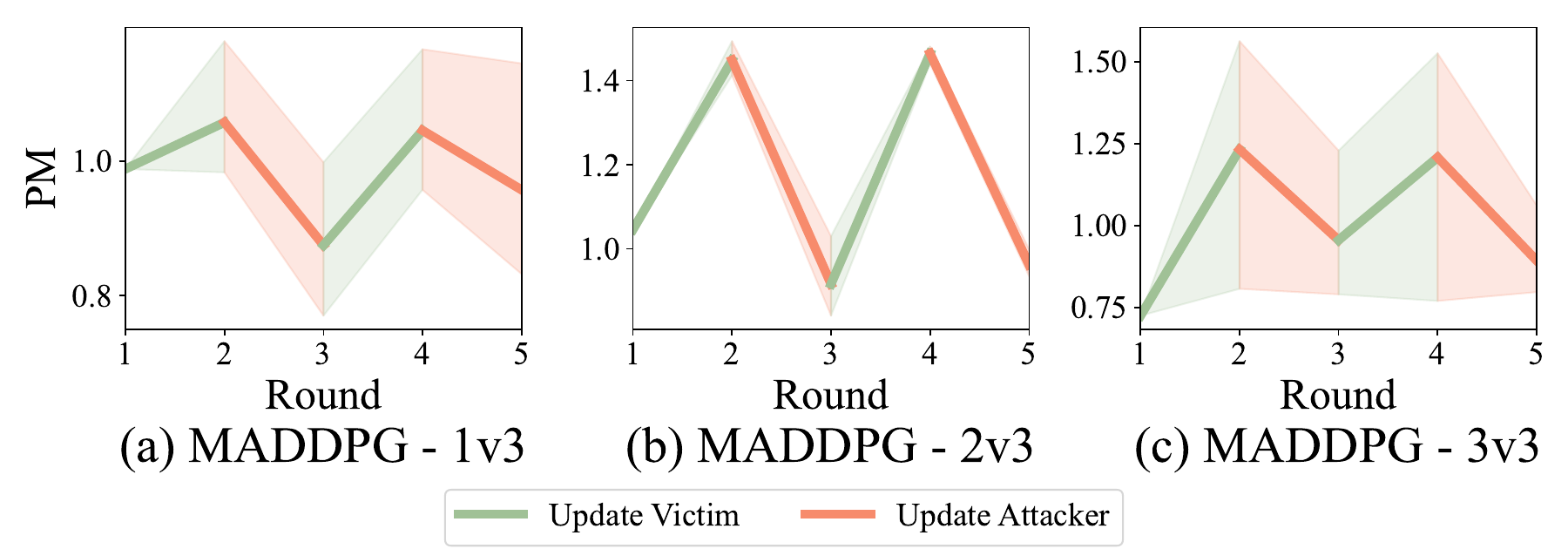}
\caption{Adversarial retraining results of 5 rounds.}
\label{fig:adv_train}
\end{figure}

\noindent \textbf{Adversarial Retraining.}
An intuitive approach involves the victim adopting adversarial retraining during the training phase~\cite{guo2023patrol}.
This allows the victim to identify and address vulnerabilities in its policy continuously.
However, as shown in~\autoref{fig:adv_train}, this does not render the adversarial policy ineffective.
This is because there is no theoretical evidence indicates that the victim's MARL policy, after adversarial retraining, can approach a Nash equilibrium more closely.
Thus, the exploitable space persists, albeit potentially shifting with adversarial retraining.
Moreover, as previously mentioned, finding or approximating a Nash equilibrium in a multi-agent competition is at least as tricky as PPAD-complete.

\noindent \textbf{Policy Ensemble.}
While it is challenging to eliminate the threat posed by adversarial policies through training alone, victims could mitigate this threat by adjusting MARL's deployment strategy.
One such strategy involves adopting a policy ensemble approach, where the victim prepares a set of policies and consistently selects policies from this ensemble for deployment.
Intuitively, this may prevent the attacker from adapting to a specific policy.

The results on the left side of~\autoref{tab:ensemble} indicate that if attackers have access to all policies, the policy ensemble shows limited defensive effect.
For instance, the entry in the first row and first column, -0.07, denotes a decrease of 0.07\% in the effectiveness of adversarial policies, which is almost negligible (note that -100\% corresponds to the failure of the adversarial policy).
This implies that the assumption of freezing the victim's policy in~\autoref{sec:Threat_Model} can be relaxed.

If attackers have access to only a subset of the victim's policies (33\%), there is a certain degree of reduction in the effectiveness of adversarial policies (the results on the right side of~\autoref{tab:ensemble}).
The insight derived from this is that the victim could periodically update the ensemble pool to prevent the attacker from adapting to all policies.
The victim may also consider increasing the diversity of the policies in the pool~\cite{fu2023iteratively, henderson2018deep}, making their weaknesses significantly different, which might lead to fluctuations in adversarial policies, preventing them from converging.
Policy ensemble can be coupled with a dynamic switching mechanism to enhance the defense approach further, wherein the switching time or policy selection is dynamically changed~\cite{10045795, 9565821, jiang2023text}.

\begin{figure}[tb]
\centering
\includegraphics[height=1.35 in]{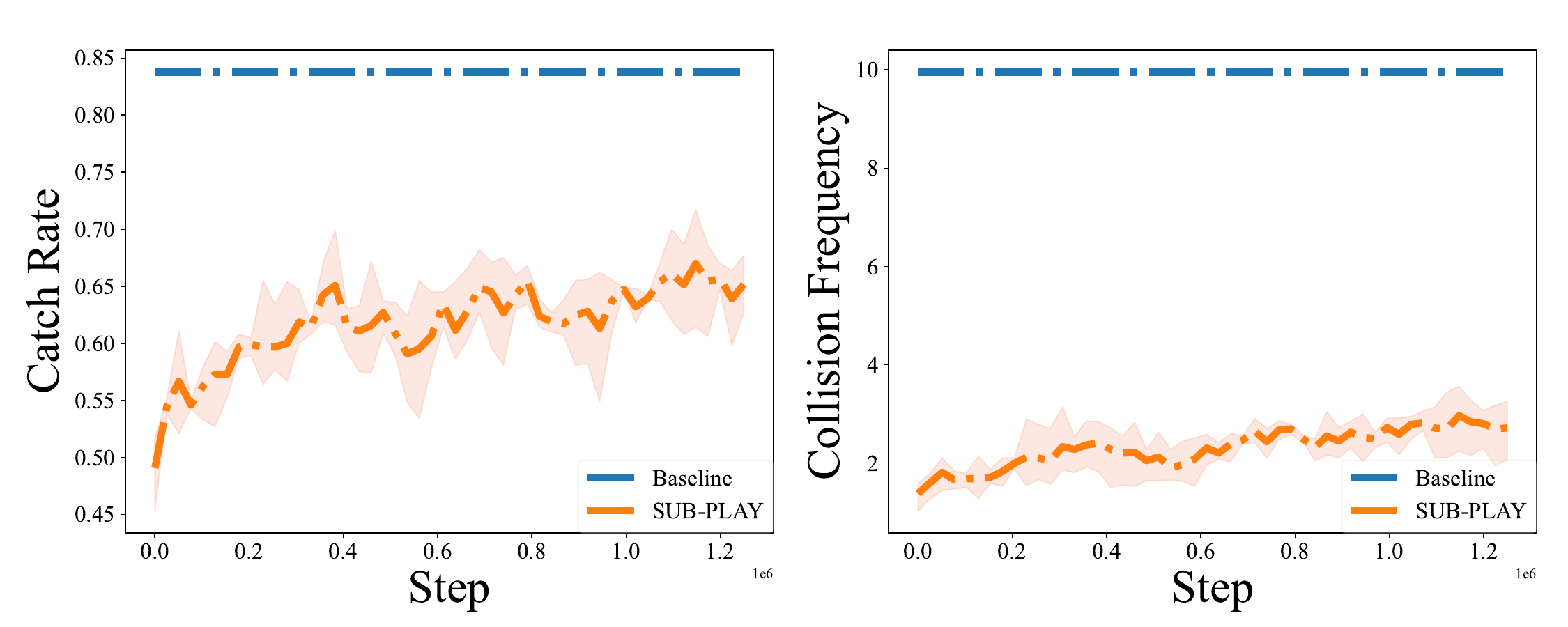}
\caption{The attack performance varies with the fine-tuning of the victim's policy.}
\label{fig:fine_tune}
\end{figure}

\noindent \textbf{Fine-tuning.}
Continual fine-tuning during deployment may also prevent the attacker from adapting to a specific victim, offering lower training costs than policy ensemble.
However, the limited defensive effectiveness of fine-tuning, as shown in~\autoref{fig:fine_tune}, suggests that the distance between policies before and after fine-tuning remains close.

\begin{table}
\scriptsize
\renewcommand\arraystretch{1.2}
\caption{The difference in training time between adversarial policies and victims (s).}
\label{tab:training_time}
\centering
\begin{tabular}{|c|ccccc|c|}\hline
\multirow{2}*{Methods} & \multicolumn{5}{c|}{Scenarios} & \multirow{2}*{Average} \\ \cline{2-6}
\multirow{2}{*}{} & 1v3 & 2v3 & 3v3 & 2v2 & 4v2 & \multirow{2}{*}{} \\ \hline
\multicolumn{7}{|c|}{\textbf{DDPG}}\\
\hline
Victim & 53240 & 68101 & 80800 & 53314 & 70364 & 65164 \\ \hline
\emph{Victim-play} & 625 & 766 & 982 & 738 & 1288 & 880 \\ \hline
\emph{SUB-PLAY} & 929 & 1293 & 1732 & 1173 & 2519 & 1529 \\ \hline
\multicolumn{7}{|c|}{\textbf{MADDPG}}\\ \hline
Victim & 92048 & 106458 & 147257 & 98678 & 131235 & 115135 \\ \hline
\emph{Victim-play} & 828 & 1172 & 1549 & 1115 & 2201 & 1373 \\ \hline
\emph{SUB-PLAY} & 1577 & 2556 & 3682 & 2304 & 3785 & 2781 \\ \hline
\end{tabular}
\end{table}

\section{Discussion}
\label{sec:Discussion}

\noindent \textbf{Emphasizing Deployment Details.}
We not only introduce \emph{SUB-PLAY} but also reveal that even with partial observations, adversarial policies induce the failure of MARL.
Since mitigating the threat of adversarial policies through improvements in the training framework is challenging, we propose that defenders prioritize the deployment details of MARL rather than solely focusing on enhancing algorithm performance.
Additionally, \emph{SUB-PLAY} can serve as a method to measure the lower bound of MARL performance in adversarial scenarios.

\noindent \textbf{Limitation and Future Work.}
\emph{SUB-PLAY} still has some limitations.
(1) Due to the scarcity of environments that facilitate multi-agent competitive settings with partial observability, our testing is limited to two environments.
However, we extensively evaluate \emph{SUB-PLAY} in various settings, including different types of partial observability limitations, multiple scenarios with varying numbers of agents, and two MARL architectures.
(2) Although no additional interaction is required, the training costs incurred by \emph{SUB-PLAY} are directly proportional to the number of subgames.
Thus, a trade-off exists between training costs and attack performance, which still needs to be addressed.
(3) The current method assumes that the attacker engages in multiple interactions with the victim. We plan to adopt offline RL techniques~\cite{du2023orl,dai2024mamba} to relax this setup.

\noindent \textbf{Real-world Scenarios.}
Agents' restricted perception capabilities and observation permissions give rise to numerous partially observable scenarios in real-world settings.
For instance, the anticipated applications of drone swarms and robots in MAS encompass encirclement systems~\cite{hafez2015solving}, security systems~\cite{hell2019drone}, strategic maneuvers~\cite{fernandez2021multi}, and human-robot teams~\cite{de2020towards}.
Nonetheless, their environmental perception is confined by the deployed sensors.
For example, the 4D LiDAR L1 on Unitree Go2 has a scanning distance of 30 meters~\cite{Go2}, leading to partially observable phenomena when the targets exceed this range or become obstructed.

\noindent \textbf{Potential Damages.}
The potential damages of \emph{SUB-PLAY} include attaining targeted victories or illicit profits. For instance, these could involve defeating specific opponents in RoboMaster~\cite{Robomaster}, exploiting strategic vulnerabilities in poker AI like Dou Dizhu~\cite{you2020combinatorial} to gain illegal profits online (given the prevalence of AI in online poker), or bypassing security MAS to jeopardize property and personal safety.
To address the potential damages posed by \emph{SUB-PLAY}, we explore potential defense methods in ~\autoref{sec:defenses} and provide directions for future research: Compared to costly adversarial retraining and limited defensive performance of fine-tuning, deploying MARL in the form of policy ensembles and increasing the diversity of the policy pool is a more practical and effective approach.

\begin{table}
\scriptsize
\renewcommand\arraystretch{1.2}
\caption{The defensive effectiveness of policy ensemble, with values given in percentage (\%).}
\label{tab:ensemble}
\centering
\begin{tabular}{|c|c|c|c|c|c|c|}\hline
Access & \multicolumn{3}{c|}{100\%} & \multicolumn{3}{c|}{33\%}\\ \hline
Scenarios & 1v3 & 2v3 & 3v3 & 1v3 & 2v3 & 3v3\\
\hline
\multicolumn{7}{|c|}{\textbf{Uncertainty}}\\
\hline
0.00 & -0.07 & +0.02 & -0.04 & -2.74 & +4.09 & -0.89 \\ \hline
0.25 & +0.02 & -0.25 & +0.10 & -9.86 & -13.58 & -12.45 \\ \hline
0.50 & +0.00 & -0.02 & +0.08 & -9.68 & -9.14 & -17.01 \\ \hline
0.75 & -0.01 & -0.07 & +0.04 & -15.55 & -2.56 & +2.85 \\ \hline
1.00 & +0.00 & +0.04 & +0.08 & -25.78 & -0.55 & +9.68 \\ \hline
\multicolumn{7}{|c|}{\textbf{Distance}}\\
\hline
0.5 & -0.09 & -0.15 & -0.03 & -16.17 & -7.99 & -11.98 \\ \hline
1.0 & -0.12 & -0.12 & -0.01 & -30.15 & -5.65 & +0.25 \\ \hline
1.5 & -0.29 & -0.12 & -0.02 & -20.24 & -9.36 & -32.69 \\ \hline
2.0 & -0.13 & -0.28 & +0.14 & -16.01 & -20.51 & -43.39 \\ \hline
\multicolumn{7}{|c|}{\textbf{Region}}\\
\hline
1 & -0.08 & -0.24 & +0.00 & -7.99 & -37.44 & -17.94 \\ \hline
\end{tabular}
\end{table}

\section{Related Work}
\label{sec:Related Work}

\subsection{RL Security}
A substantial body of research is leveraging RL to achieve specific security objectives ~\cite{wang2021reinforcement, gohil2022attrition, yu2023airs, maiti2023targeted, xia2023rlid}.
Nevertheless, the security dimensions inherent to RL need to be more adequately addressed.
This section thoroughly examines security research concerning the three fundamental components of RL.

\noindent \textbf{Reward Manipulation.}
Unlike the modification of labels in deep learning~\cite{wang2019neural, yao2019latent}, RL introduces backdoor attacks via reward manipulation~\cite{ma2019policy, kiourti2020trojdrl}.
Zhang \emph{et al.} \cite{zhang2020adaptive} developed a dynamic reward-poisoning attack targeting online RL applications, while Chen \emph{et al.}~\cite{chen2022marnet} extended the concept of backdoor attacks to cooperative MASs.
Wang \emph{et al.}~\cite{wang2021backdoorl} introduced a unique RL-specific paradigm for backdoor attacks, where the attacker trains a benign policy and a trojan policy, merging them into a backdoor policy by behavior cloning.
Guo \emph{et al.}~\cite{guo2023policycleanse} discovered a pseudo-trigger space that can trigger RL backdoors.
In response, they proposed PolicyCleanse to perform model detection and backdoor mitigation.

\noindent \textbf{State Manipulation.}
Inspired by adversarial examples~\cite{carlini2017towards, yu2020cloudleak, li2020advpulse}, attackers in RL can disrupt the victim by perturbing the environment state.
Huang \emph{et al.}~\cite{huang2017adversarial} applied FGSM~\cite{goodfellow2015explaining} to DRL and launched an adversarial attack on the DQN policy in Atari games~\cite{mnih2015human}.
Behzadan \emph{et al.}~\cite{behzadan2017vulnerability} introduced a policy induction attack, where the attacker determines the victim's actions based on a pre-trained target policy and perturbs the states by FGSM and JSMA~\cite{papernot2016limitations}.
Sun \emph{et al.}~\cite{sun2020stealthy} proposed a white-box attack called the critical point attack, which strategically explores state-action combinations to identify points with high payoff and inject subtle perturbations during the victim's deployment phase.

\noindent \textbf{Action Manipulation.}
The agent's action determines the agent-environment boundary, allowing attackers to launch attacks by manipulating it.
Lee \emph{et al.}~\cite{lee2020spatiotemporally} proposed two victim manipulation attacks: the myopic action-space attack injects action perturbations based on current observations, while the look-ahead action-space attack considers future steps to maximize the attack's impact.
However, directly manipulating the victim's actions is impractical.

In contrast, adversarial policies only need to control the attacker's action.
Gleave \emph{et al.}~\cite{gleave2020adversarial} pioneered \emph{Victim-play}, which involves manipulating the adversary's actions to induce suboptimal decisions from a fixed RL model during deployment.
Wu \emph{et al.}~\cite{wu2021adversarial} incorporated explainable AI techniques into adversarial policy generation, enhancing the stealthiness by launching attacks only when the victim pays attention to them.
Guo \emph{et al.}~\cite{guo2021adversarial} extended \emph{Victim-play} from zero-sum to general-sum environments, revealing its potential in assessing the fairness of competitions or systems.
Wang \emph{et al.}~\cite{wang2023adversarial} explored adversarial policies in discrete action scenarios and achieved success against superhuman-level Go AIs, demonstrating that \emph{near-Nash} or $\epsilon\emph{-equilibrium}$ policies are exploitable.
Guo \emph{et al.}~\cite{guo2023patrol} attempted to mitigate the potential threat of adversarial policies from a training perspective and introduced a provable defense called PATROL.
The aim is to bring the victim closer to a Nash equilibrium in a two-player competition.
Liu \emph{et al.}~\cite{liu2023rethinking} explored adversarial policy attacks in scenarios where attackers only have partial control over the adversary and proposed adversarial training with two timescales to mitigate the threats posed by adversarial policies.

\subsection{Partial Observability in RL}
Partial observability limits the attacker's access to complete environmental information. To address this, existing research suggests two potential methods for the attacker.

\noindent \textbf{Inference.} Inference entails using available observations and prior knowledge to complete unobserved content, including environmental~\cite{yang2021program, xu2022side} and agent inference~\cite{papoudakis2021agent, gu2022online}.
(1) Environmental Inference:
Partial state information is used to infer the global environment in the spatial dimension.
Yang \emph{et al.}~\cite{yang2021program} proposed a supervised learning-based hallucinator for inferring the environment from current observations, effective in static environments but potentially less suitable for highly dynamic competitions.
(2) Agent Inference:
Historical interaction is used to infer the unobservable agents in the temporal dimension.
Papoudakis \emph{et al.}~\cite{papoudakis2021agent} proposed constructing policies for all agents through representation learning.
During deployment, the policies and local observations of the controlled agent are utilized to infer the invisible agents.
However, this method relies on the victim's policy knowledge and is unsuitable for black-box or competitive environments.

\noindent \textbf{Generalization.}
Enhancing the generalization capability of agents can effectively adapt to the environment's diversity, dynamics, and unpredictability~\cite{kirk2023survey}.
Intuitively, this also applies to policy improvements in partially observable scenarios.
Ghosh \emph{et al.}~\cite{ghosh2021generalization} demonstrated that partitioning a partially observable task into multiple subtasks can effectively improve the performance of RL policies.
\emph{SUB-PLAY} draws on the insight of generalization-based approaches to address the partially observable problem in multi-agent competitive environments, as it applies to dynamic environments and does not rely on additional victim information.

\section{Conclusion}
\label{sec:Conclusion}

This paper proposes \emph{SUB-PLAY}, a novel black-box attack framework in partially observable multi-agent competitive environments.
The effectiveness of \emph{SUB-PLAY} in enhancing the attack performance of adversarial policies is showcased through divide-and-conquer strategies and transition dissemination, as evidenced by extensive evaluations conducted across various partially observable limitations and MARL algorithms.
Moreover, we examine three potential defense strategies to mitigate the risks associated with \emph{SUB-PLAY}. 
The evaluation results indicate that policy ensemble is more effective than adversarial retraining and fine-tuning. 
Future investigations can concentrate on enhancing the diversity of policy pools and implementing mechanisms for dynamic policy switching.

\section*{Acknowledgment}
We sincerely appreciate the insightful comments from our shepherd and the anonymous reviewers. 
We would like to extend our gratitude to Xuhong Zhang, Chenghui Shi, Jiang Yi, Yuyou Gan, Guang Yang and Jiawen Wan for their valuable feedback. 
This work was partially supported by the National Key Research and Development Program of China under grant number 2022YFB3102100.

\bibliographystyle{ACM-Reference-Format}
\bibliography{subplay}

\appendix

\section{Exploitable Ceiling}
\label{app:Exploitable Ceiling}

In a ZS-POSG, an adversarial policy can be interpreted as an exploitative strategy akin to the strategies used in competitions like Texas Hold'em poker \cite{brown2019superhuman}.
Exploitative strategies capitalize on opponents' vulnerabilities or tendencies, diverging from strict adherence to optimal or equilibrium strategy.
Additionally, it's essential to note that real-world competitions, such as military conflicts, often lack the same level of well-defined rules and fairness seen in games like Texas Hold'em poker.
This implies that one of the MAS, whether the adversary or the victim, can benefit from the vulnerabilities in the competition rules.
Therefore, the attack performance of an adversarial policy relies on the victim's policy and the competition rule, with an upper bound referred to as the exploitable ceiling.

\noindent \textbf{Definition 3.}
\textit{Exploitable ceiling refers to the limit of the performance degradation of the victim's policy in a specific competitive environment, which is formalized as}
$$
\mathcal{C} = \mathcal{C} (\pi_V) + \mathcal{C} (e),
$$
\textit{where $\mathcal{C} (\pi_V)$ denotes the exploitable ceiling of the victim's policy and $\mathcal{C} (e)$ denotes the exploitable ceiling of a two-team competitive environment.}

The specific measurement of the exploitable ceiling could be approached from both the perspectives of the victim and the adversary. Considering the victim's deployed policy $\pi_V$ and assuming that the victim possesses a Nash equilibrium policy $\pi_V^*$ in the constructed POSG, the exploitable ceiling can be defined as the victim's regret:
$$
\mathcal{C} = \max\limits_{\pi_V^* \in \mathbb{P}_V}R(\pi_V, \pi_V^*),
$$
where $\pi_V^*$ denotes a Nash equilibrium policy of the victim in the environment. $R(\pi_V, \pi_V^*)$ denotes the expected regret of policy $\pi_V$ against $\pi_V^*$, which is defined as:
$$
R(\pi_V, \pi_V^*) = \mathbb{E}[\sum_{t=0}^T (r_t (\pi_V, b_t) - r_t (\pi_V^*, b_t^*))],
$$
where $r_t$ is the reward received by the victim at time $t$, $b_t$ is the action taken by policy $\pi_V$ at time $t$, $b_t^*$ is the action taken by $\pi_V^*$ at time $t$, and $T$ is the time horizon.

However, the existence of a Nash equilibrium is not guaranteed in a ZS-POSG due to an expanded policy space and challenges from incomplete information. In such cases, the exploitable ceiling can be defined as the adversary's best response to the policy $\pi_V$:
$$
\mathcal{C} = \max\limits_{\pi_A \in \mathbb{P}_A}\mathbb{E}[u(\pi_A', \pi_V) - u(\pi_A, \pi_V)],
$$
where $\pi_A'$ denotes the adversary's initial policy. $u(\pi_A', \pi_V)$ is the adversary's utility function, denoting the expected accumulated reward when the adversary adopts $\pi_A'$ to compete with $\pi_V$. Similarly, $u(\pi_A, \pi_V)$ denotes the expected accumulated reward when the adversary adopts an arbitrary $\pi_A \in \mathbb{P}_A$ to compete with $\pi_V$.

\section{Proof of Proposition 1}
\label{app:a}

\noindent \textbf{Proposition 1.}
\textit{In a zero-sum partially observable stochastic game, if the victim keeps a fixed joint policy $\pi_V$, the state transition of the environment is solely dependent on the adversary's joint policy $\pi_A$.}

\noindent \textbf{Proof.} 
Given the fixed joint policy $\pi_V$ of $Victim$, the probability distribution of generating a trajectory $\tau$ can be factorized as:

$$
\begin{aligned} 
P(\tau) &= P(s_0) \prod_{t=0}^{T-1} P(s_{t+1} \vert{} s_t, a_t, b_t) \\ 
&\quad \cdot P(o_{t+1} \vert{} s_{t+1}, a_t, b_t) \pi_V(b_t \vert{} s_t) \pi_A(a_t \vert{} o_t). 
\end{aligned}
$$

During the training of the adversarial policy, the initial state distribution $P(s_0)$, the state transition probability $P(s_{t+1} \vert{} s_t, a_t, b_t)$, the observation emission probability $P(o_{t+1} \vert{} s_{t+1}, a_t, b_t)$, and $Victim$'s policy $\pi_V(b_t \vert{} s_t)$ remain stationary. 
Therefore, we can group these independent components into a fixed environmental trajectory operator $\mathcal{P}_{\text{env}, V}(\tau)$:
$$
\mathcal{C}_{\text{env}, V}(\tau) = P(s_0) \prod_{t=0}^{T-1} P(s_{t+1} \vert{} s_t, a_t, b_t) P(o_{t+1} \vert{} s_{t+1}, a_t, b_t) \pi_V(b_t \vert{} s_t).
$$
Consequently, the trajectory probability simplifies to:
$$
P(\tau) = \mathcal{C}_{\text{env}, V}(\tau) \prod_{t=0}^{T-1} \pi_A(a_t \vert{} o_t).
$$

Furthermore, the state-value function of $Adversary$ is defined as the expected cumulative return over the trajectory distribution:
$$
V(\pi_A) = \mathbb{E}_{\tau \sim P} [\mathcal{R}(\tau)] = \sum_{\tau} \left( \mathcal{C}_{\text{env}, V}(\tau) \prod_{t=0}^{T-1} \pi_A(a_t \vert{} o_t) \right) \mathcal{R}(\tau).
$$

In summary, under the premise that the joint policy $\pi_V$ of $Victim$ is fixed, all variations in the trajectory distribution $P(\tau)$ and the state-value function $V(\pi_A)$ are solely determined by $Adversary$'s current joint policy.
\hfill $\blacksquare$

\begin{figure}[t]
\centering
\includegraphics[height=2.7 in]{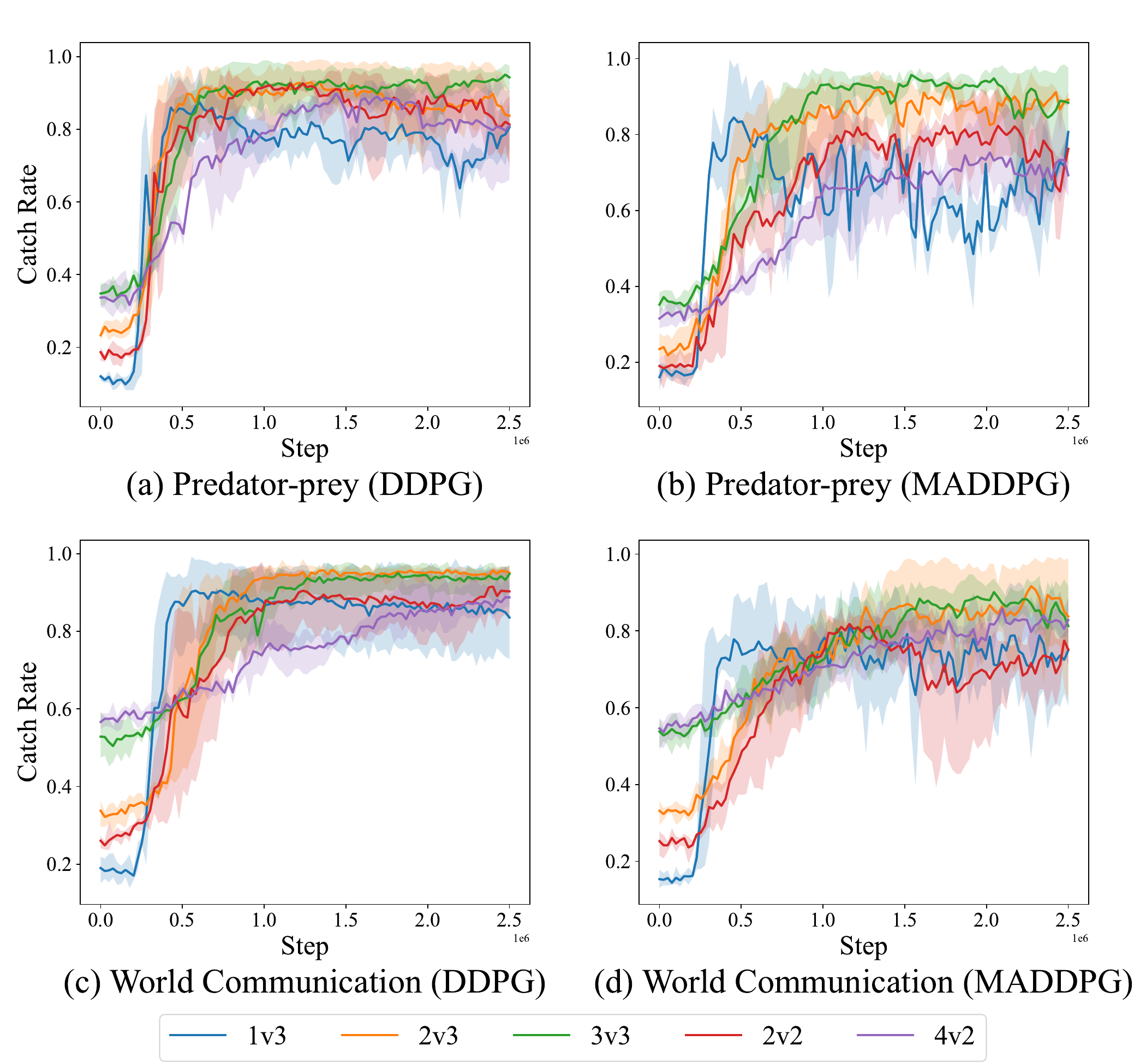}
\caption{The training curves of victims in diverse scenarios.}
\label{fig:curve_train}
\end{figure}

\begin{figure}
\centering
\includegraphics[height=2.6 in]{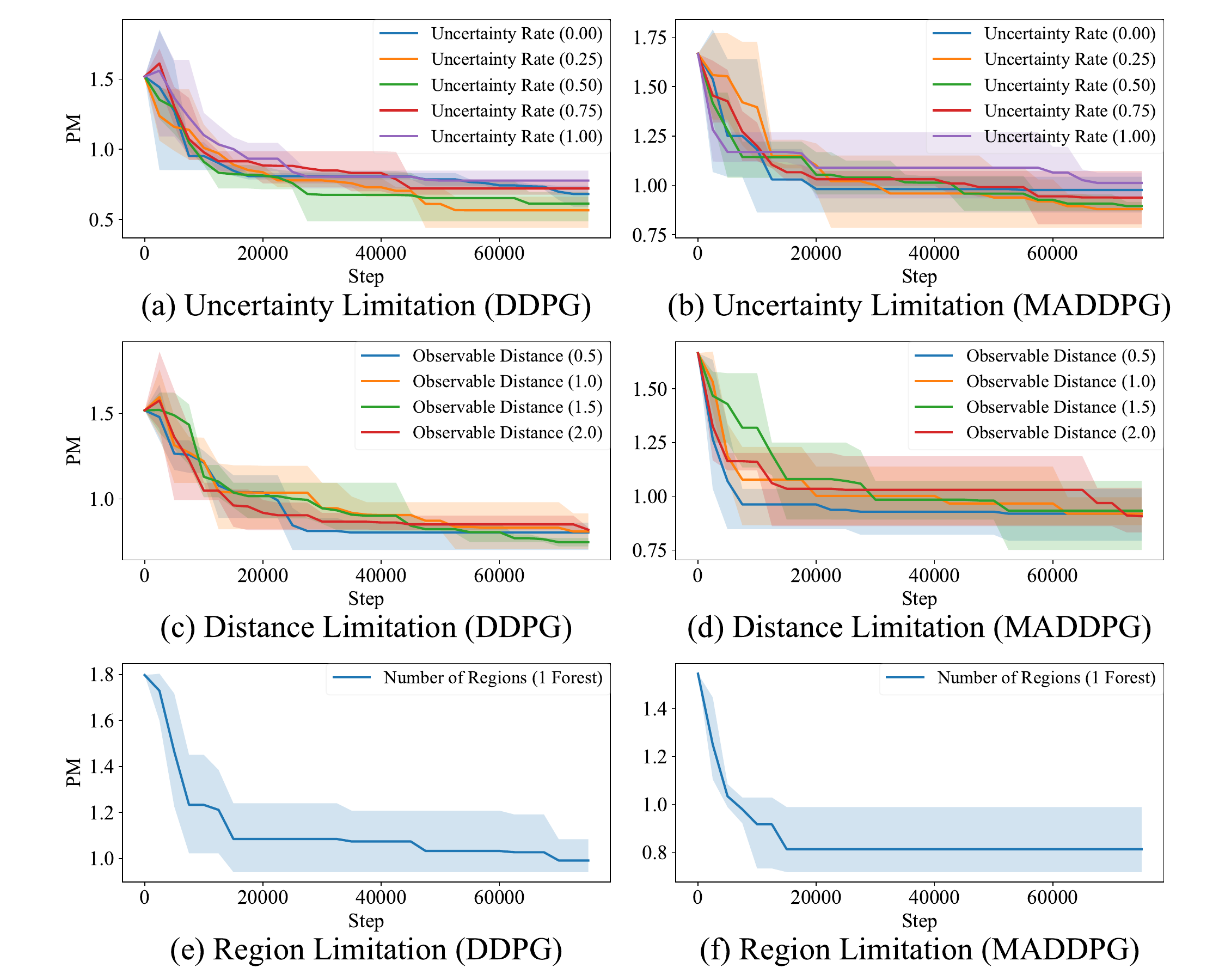}
\caption{The variation in attack performance of adversarial policies.}
\label{fig:curve_attack}
\end{figure}

\begin{algorithm}
\caption{Subpolicy Training.}
\label{alg:Subpolicy_Training}
\begin{algorithmic}[1]

\STATE \textbf{Input:}
$Adversary$'s latest subpolicies $\{ \pi_{\alpha_k}^i \}_{i \in \mathcal{M}, k \in \mathcal{K}}$, replay buffers $\{ \mathcal{E}_k^i \}_{i \in \mathcal{M}, k \in \mathcal{K}}$, and recorded test performance $\{ PM_k^i \}_{i \in \mathcal{M}, k \in \mathcal{K}}$.
\STATE \textbf{for} $i = 1, 2, ..., M$ \textbf{do}
\STATE \quad \textbf{for} $k = 0, 1, ..., \emph{Sub} - 1$ \textbf{do}
\STATE \qquad Randomly select a batch of transitions from $\mathcal{E}_k^i$.
\STATE \qquad Obtain $\overline{\pi}_{\alpha_k}^i$ through MARL training.
\STATE \qquad Test the performance of $\overline{\pi}_{\alpha_k}^i$ on each metric and record it as $\{ \eta_{k_0}^i, \eta_{k_1}^i,..., \eta_{k_L}^i \}$.
\STATE \qquad Calculate $\overline{PM}_k^i$ by \autoref{eq:pm}.
\STATE \qquad \textbf{if} $\overline{PM}_k^i$ outperforms $PM_k^i$ \textbf{then}
\STATE \quad \qquad $ \pi_{\alpha_k}^i \gets \overline{\pi}_{\alpha_k}^i $, $ PM_k^i \gets \overline{PM}_k^i $
\STATE \qquad \textbf{end if}
\STATE \quad \textbf{end for}
\STATE \textbf{end for}
\STATE \textbf{Output:}
Updated $\{ \pi_{\alpha_k}^i \}_{k \in \mathcal{K}, i \in \mathcal{M}}$ and $\{ PM_k^i \}_{k \in \mathcal{K}, i \in \mathcal{M}}$

\end{algorithmic}
\end{algorithm}

\begin{algorithm}
\caption{Policy Combination (Distributed MARL Version).}
\label{alg:Policy Combination}
\begin{algorithmic}[1]

\STATE \textbf{Input:}
$Adversary$'s joint observation $(o_1, o_2,..., o_M)$, the constructed subgames $\{ \mathcal{G}_{\alpha_k} \}_{k \in \mathcal{K}}$, the set $\{ \pi_{\alpha_k}^i \}_{k \in \mathcal{K}, i \in \mathcal{M}}$ consisting of all subpolicies.
\STATE \textbf{for} $i = 1, 2, ..., M $ \textbf{do}
\STATE \quad \textbf{for} $k = 0, 1, ..., \emph{Sub} - 1 $ \textbf{do}
\STATE \qquad \textbf{if} $o_i \in \Omega_k^i $ \textbf{then}
\STATE \quad \qquad $ a_i \gets \pi_{\alpha_k}^i (o_i) $
\STATE \qquad \textbf{end if}
\STATE \quad \textbf{end for}
\STATE \textbf{end for}
\STATE \textbf{Output:}
$Adversary$'s joint action $(a_1, a_2,..., a_M)$.

\end{algorithmic}
\end{algorithm}

\begin{algorithm}
\caption{Partially Observable Implementation.}
\label{alg:Partially Observable Implementation}
\begin{algorithmic}[1]

\STATE \textbf{Input:}
The joint state provided by the environment $(s_1, s_2,..., s_M)$, the partially observable limitation $\mathcal{L}$.
\STATE \textbf{for} $i = 1, 2, ..., M $ \textbf{do}
\STATE \quad \textbf{if} $\mathcal{L} = $ Uncertainty Limitation \textbf{then}
\STATE \qquad Generate a mask vector $m_u$ based on the uncertainty rate.
\STATE \qquad $ o_i \gets s_i \odot m_u $
\STATE \quad \textbf{end if}
\STATE \quad \textbf{if} $\mathcal{L} = $ Distance Limitation \textbf{then}
\STATE \qquad Generate a mask vector $m_d$ based on the observable distance.
\STATE \qquad $ o_i \gets s_i \odot m_d $
\STATE \quad \textbf{end if}
\STATE \quad \textbf{if} $\mathcal{L} = $ Region Limitation \textbf{then}
\STATE \qquad $ o_i \gets s_i$
\STATE \quad \textbf{end if}
\STATE \textbf{end for}
\STATE \textbf{Output:}
Modified joint observation $(o_1, o_2, ..., o_M)$.

\end{algorithmic}
\end{algorithm}

\begin{figure}
\centering
\includegraphics[height=2.0 in]{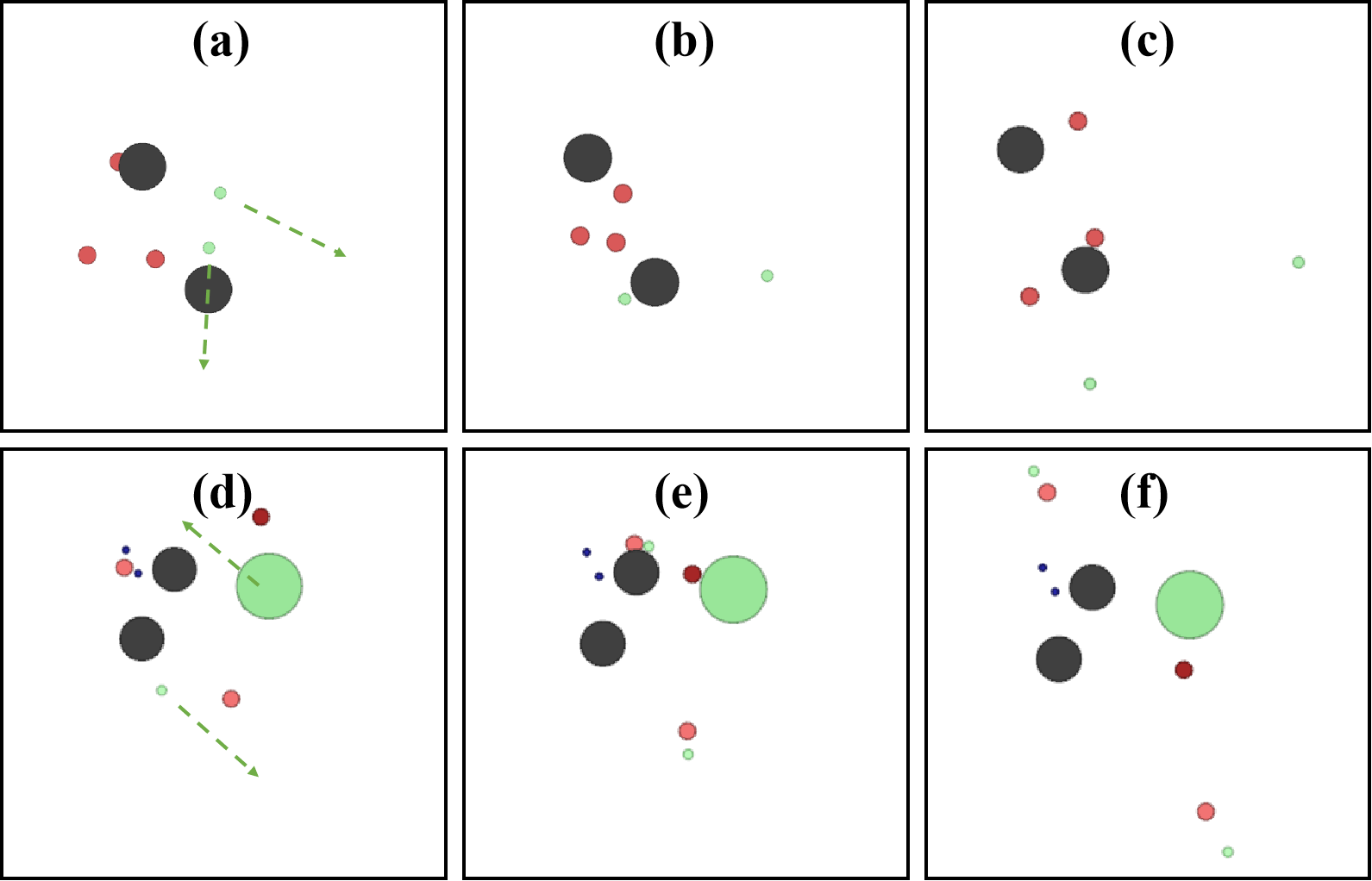}
\caption{The visualization of adversarial policies. (a)-(c) are visualizations of one episode in Predator-prey. (d)-(f) are visualizations of one episode in World Communication.}
\label{fig:visualization}
\end{figure}

\begin{figure}
\centering
\includegraphics[height=4.8 in]{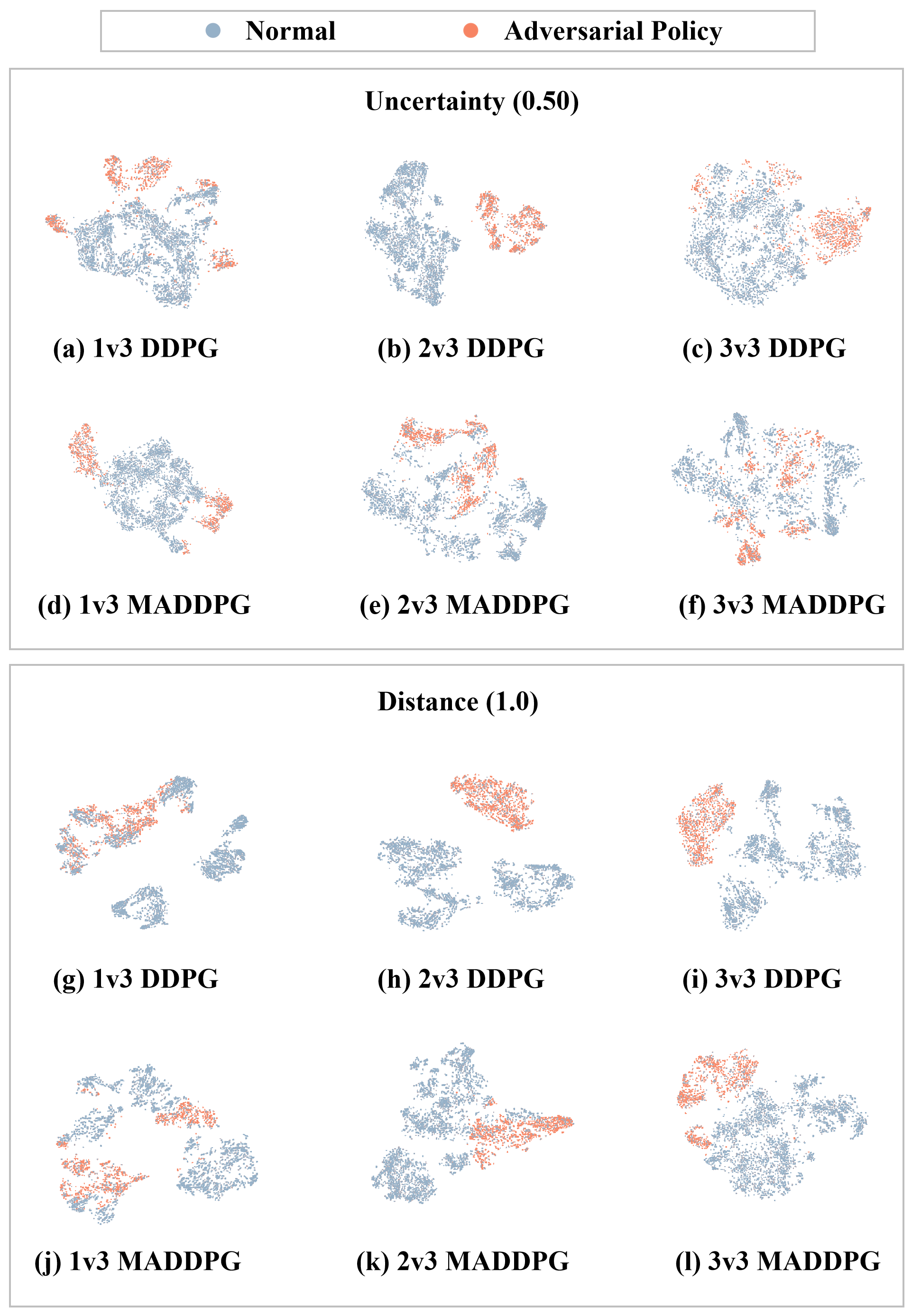}
\caption{Additional t-SNE visualizations. Model fitted with a perplexity of 250 to activations from 5000 timesteps against each opponent.}
\label{fig:tsne2}
\end{figure}

\section{Additional Implementation Details}
\label{app:Additional Evaluation Details}

\noindent \textbf{Victim Training. }
The victims are derived from \emph{Self-play}, which is a general framework in RL competitive environments \cite{bansal2018emergent}.
Each victim is equipped with a replay buffer of size 200,000 and a batch size of 1024.
Throughout the training phase, the victims undertake 100,000 episodes, which corresponds to 2,500,000 steps.
To account for variations, we train three victims for every scenario, each initialized with a different random seed.
The training progress of these victims is illustrated in~\autoref{fig:curve_train}.

\noindent \textbf{Adversarial Policy Generation. }
During the attack phase, each adversary agent utilizes a replay buffer with a size of 512 and a batch size of 512.
Similar to the victim agents, we train three separate adversarial policies with different random seeds for each victim.
The training curves of these adversarial policies can be observed in~\autoref{fig:curve_attack}.
Notably, the victim agents require an average of 700,000 steps to converge, whereas the adversarial policies achieve convergence in only approximately 20,000 steps (compare~\autoref{fig:curve_train} and~\autoref{fig:curve_attack}).
These results indicate a notable disparity in the convergence rates between the victim and adversary.

\begin{figure*}
\centering
\includegraphics[width=\textwidth]{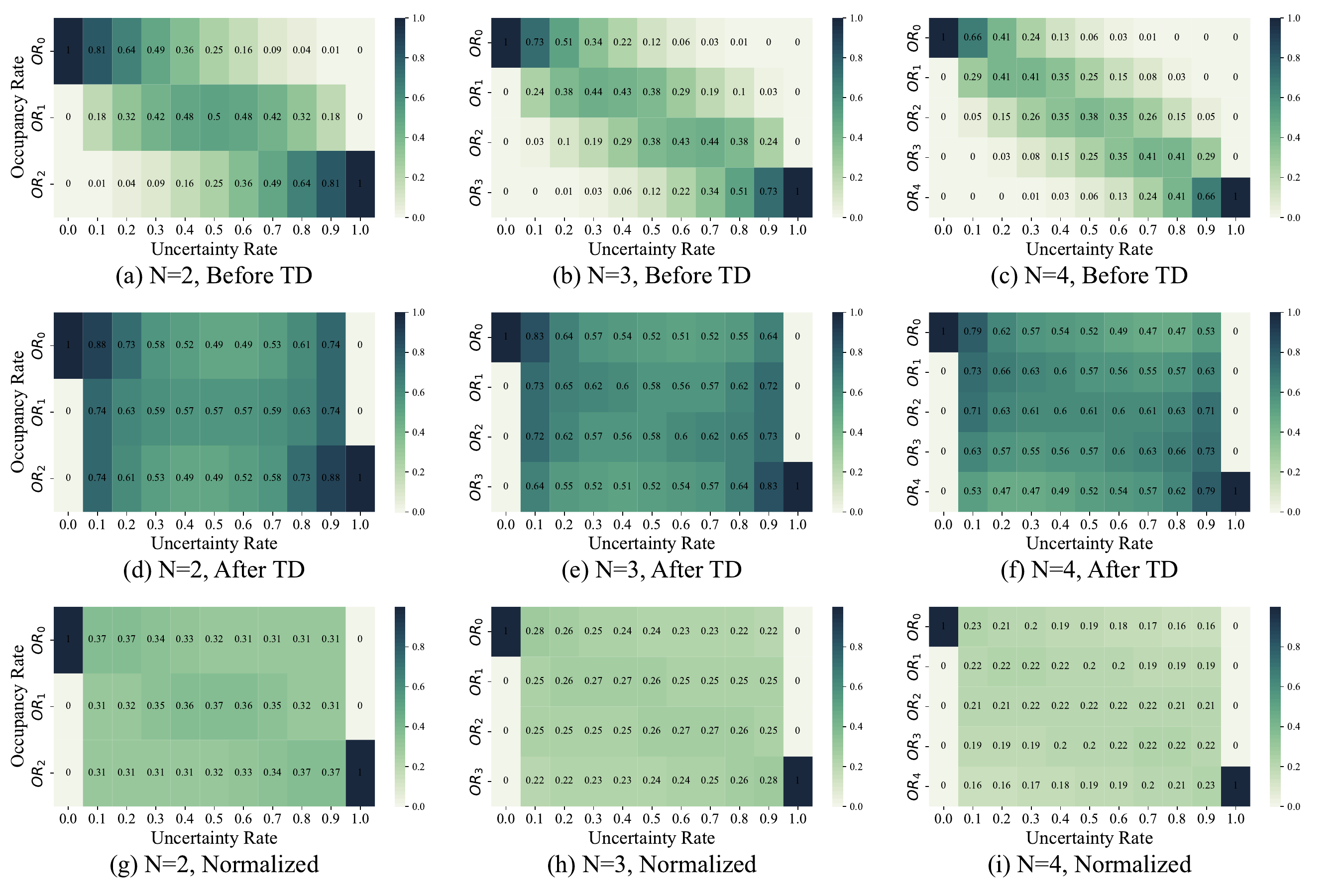}
\caption{The occupancy rate of transitions for each replay buffer in different environments under uncertainty limitations. (a)-(c) denote the cases without transition dissemination; (d)-(f) denote the cases with transition dissemination; (g)-(i) denote the normalized result of occupancy rate after transition dissemination.}
\label{fig:heatmap}
\end{figure*}

\section{Additional Comparison}
\label{app:Additional Comparison}

We compare \emph{SUB-PLAY} with two additional variants of \emph{Victim-play} (\emph{APL}~\cite{guo2021adversarial}, \emph{Victim-curr}~\cite{wang2023adversarial}), further validating \emph{SUB-PLAY}'s attack performance in partially observable environments.

In addition to the vanilla version~\cite{gleave2020adversarial}, \emph{Victim-play} also has three additional variants.
Wu \emph{et al.}~\cite{wu2021adversarial} incorporated explainable AI techniques into adversarial policy generation, enhancing stealthiness by launching attacks only when the victim pays attention to them.
However, this method only considers the case where the victim is a single agent and cannot be directly applied to scenarios where the victim is a MAS, as it is unclear how to integrate the attention of multiple agents.
Therefore, we compare the remaining two variants, \emph{APL} and \emph{Victim-curr}.
Although they are proposed for general-sum game and sparse reward scenarios in two-player competitions, with minor adjustments, they can be deployed in multi-agent scenarios.

\noindent \textbf{Implementation details.}
In \emph{Victim-play}, the attacker aims to maximize the adversary's reward, while \emph{APL}'s intuition changes the objective to maximize the difference in rewards between the adversary and the victim.
In \emph{APL}, the attacker approximates the victim's reward.
In our implementation, the attacker can obtain the reward information from the victim at each time step (gray-box setting), thereby enhancing the attack capability of \emph{APL}.

In \emph{Victim-curr}, the attacker is granted access to multiple checkpoints of the victim's training process from the beginning (gray-box setting) and utilizes a curriculum learning approach.
The curriculum involves the adversary initially interacting with early versions of the victim and updating until achieving over a 55\% win rate in the game of Go, at which point the victim's updated version is introduced.
This process repeats until the adversarial policy can defeat the deployed victim.
In our implementation, the attacker can access 6 out of 100 victim checkpoints (numbered 1, 20, 40, 60, 80, 100).
Due to the asymmetry of the Predator-prey and World Communication environments and the absence of a win rate metric, the attacker is set to replace checkpoints at a fixed time interval.

\noindent \textbf{Evaluation results.}
The results in ~\autoref{tab:add 1v3} -~\autoref{tab:add 4v2} indicate that the proportions of the four methods (\emph{Victim-play}, \emph{APL}, \emph{Victim-curr}, \emph{SUB-PLAY}) achieving the best attack performance under the catch rate (CR) metric are 2\%, 11\%, 22\%, and 65\% (across 100 experimental settings), respectively, while under the collision frequency (CF) metric, the proportions are 1\%, 7\%, 27\%, and 65\% (across 100 experimental settings), respectively.
Furthermore, \emph{SUB-PLAY} demonstrates significantly improved attack performance with the MADDPG algorithm and in the World Communication environment, outperforming other methods in 78\% and 80\% of scenarios, respectively.
This suggests that in larger input dimensions and more complex environments, \emph{SUB-PLAY} poses a greater threat to partially observable MASs than the other three methods.

\begin{table*}[htbp]
\scriptsize
\renewcommand\arraystretch{1.2}
\caption{The attack performance comparison in 1v3 scenarios, measured by two metrics (CR$\downarrow$/CF$\downarrow$).}
\label{tab:add 1v3}
\centering
\begin{tabular}{|c|c|cccc|cccc|}\hline
\multicolumn{2}{|c|}{\multirow{2}*{\textbf{Limitations}}} & \multicolumn{4}{c|}{DDPG} & \multicolumn{4}{c|}{MADDPG}\\ \cline{3-10}
\multicolumn{2}{|c|}{\multirow{2}{*}{}} & \emph{Victim-play} & \emph{APL} & \emph{Victim-curr} & \emph{SUB-PLAY} & \emph{Victim-play} & \emph{APL} & \emph{Victim-curr} & \emph{SUB-PLAY}  \\ \hline
\multirow{5}*{Uncertainty} & 0.00 & 0.277 / 0.803 & 0.344 / 1.148 & \textcolor{mygreen}{0.254} / \textcolor{mygreen}{0.685} & 0.265 / 0.785 & 0.353 / 1.096 & 0.351 / 1.030 & 0.264 / 0.798 & \textcolor{mygreen}{0.255} / \textcolor{mygreen}{0.756} \\
\multirow{5}{*}{} & 0.25 & 0.390 / 1.659 & 0.503 / 3.150 & 0.259 / 0.897 & \textcolor{mygreen}{0.196} / \textcolor{mygreen}{0.623} & 0.416 / 1.333 & 0.333 / 1.016 & 0.287 / 0.857 & \textcolor{mygreen}{0.216} / \textcolor{mygreen}{0.581} \\
\multirow{5}{*}{} & 0.50 & 0.337 / 1.196 & 0.356 / 1.215 & 0.298 / \textcolor{mygreen}{0.927} & \textcolor{mygreen}{0.276} / 0.932 & 0.318 / 1.006 & 0.359 / 1.388 & 0.346 / 1.437 & \textcolor{mygreen}{0.215} / \textcolor{mygreen}{0.526} \\
\multirow{5}{*}{} & 0.75 & 0.368 / 1.250 & 0.297 / 0.782 & 0.359 / 1.255 & \textcolor{mygreen}{0.250} / \textcolor{mygreen}{0.768} & 0.322 / 0.875 & 0.401 / 1.305 & 0.358 / 1.224 & \textcolor{mygreen}{0.284} / \textcolor{mygreen}{0.843} \\
\multirow{5}{*}{} & 1.00 & 0.366 / 1.271 & \textcolor{mygreen}{0.297} / \textcolor{mygreen}{0.937} & 0.426 / 2.220 & 0.298 / 0.962 & 0.367 / 1.132 & 0.353 / 0.997 & 0.432 / 2.278 & \textcolor{mygreen}{0.236} / \textcolor{mygreen}{0.651} \\
\hline
\multirow{4}*{Distance} & 0.5 & 0.448 / 2.062 & 0.373 / 1.196 & \textcolor{mygreen}{0.354} / \textcolor{mygreen}{1.046} & 0.356 / 1.093 & 0.288 / 0.834 & 0.299 / 0.868 & 0.374 / 2.103 & \textcolor{mygreen}{0.275} / \textcolor{mygreen}{0.787} \\
\multirow{4}{*}{} & 1.0 & 0.473 / 2.675 & \textcolor{mygreen}{0.313} / 0.897 & 0.316 / \textcolor{mygreen}{0.890} & 0.328 / 1.081 & 0.441 / 1.923 & 0.252 / 0.712 & 0.431 / 1.832 & \textcolor{mygreen}{0.239} / \textcolor{mygreen}{0.672} \\
\multirow{4}{*}{} & 1.5 & 0.389 / 1.508 & 0.334 / \textcolor{mygreen}{1.031} & 0.350 / 1.183 & \textcolor{mygreen}{0.327} / 1.051 & 0.376 / 1.176 & 0.369 / 1.158 & 0.293 / 1.009 & \textcolor{mygreen}{0.241} / \textcolor{mygreen}{0.740} \\
\multirow{4}{*}{} & 2.0 & 0.420 / 1.413 & 0.340 / 1.206 & \textcolor{mygreen}{0.279} / \textcolor{mygreen}{0.823} & 0.359 / 1.207 & 0.383 / 1.588 & 0.370 / 1.179 & 0.356 / 1.560 & \textcolor{mygreen}{0.269} / \textcolor{mygreen}{0.766} \\
\hline
Region & 1 & 0.479 / 2.463 & 0.429 / 1.796 & 0.622 / 4.409 & \textcolor{mygreen}{0.356} / \textcolor{mygreen}{1.253} & 0.412 / 1.721 & 0.363 / 1.356 & 0.351 / 1.285 & \textcolor{mygreen}{0.236} / \textcolor{mygreen}{0.651} \\ \hline
\end{tabular}
\end{table*}

\begin{table*}[htbp]
\scriptsize
\renewcommand\arraystretch{1.2}
\caption{The attack performance comparison in 2v3 scenarios, measured by two metrics (CR$\downarrow$/CF$\downarrow$).}
\label{tab:add 2v3}
\centering
\begin{tabular}{|c|c|cccc|cccc|}\hline
\multicolumn{2}{|c|}{\multirow{2}*{\textbf{Limitations}}} & \multicolumn{4}{c|}{DDPG} & \multicolumn{4}{c|}{MADDPG}\\ \cline{3-10}
\multicolumn{2}{|c|}{\multirow{2}{*}{}} & \emph{Victim-play} & \emph{APL} & \emph{Victim-curr} & \emph{SUB-PLAY} & \emph{Victim-play} & \emph{APL} & \emph{Victim-curr} & \emph{SUB-PLAY}  \\ \hline
\multirow{5}*{Uncertainty} & 0.00 & 0.486 / 1.998 & 0.585 / 2.720 & \textcolor{mygreen}{0.454} / \textcolor{mygreen}{1.530} & 0.489 / 1.969 & 0.914 / 9.375 & 0.565 / 2.865 & \textcolor{mygreen}{0.501} / \textcolor{mygreen}{2.599} & 0.627 / 3.328 \\
\multirow{5}{*}{} & 0.25 & 0.566 / 2.548 & 0.574 / 2.533 & 0.456 / 1.645 & \textcolor{mygreen}{0.401} / \textcolor{mygreen}{1.574} & 0.783 / 7.823 & 0.693 / 4.458 & 0.679 / 5.684 & \textcolor{mygreen}{0.579} / \textcolor{mygreen}{3.053} \\
\multirow{5}{*}{} & 0.50 & 0.585 / 2.935 & 0.575 / 2.634 & 0.524 / \textcolor{mygreen}{1.883} & \textcolor{mygreen}{0.477} / 2.309 & 0.727 / 7.216 & 0.741 / 8.384 & 0.657 / 5.111 & \textcolor{mygreen}{0.583} / \textcolor{mygreen}{3.228} \\
\multirow{5}{*}{} & 0.75 & 0.589 / 2.798 & 0.528 / 2.285 & \textcolor{mygreen}{0.484} / \textcolor{mygreen}{1.590} & 0.502 / 1.928 & 0.627 / 3.826 & 0.769 / 6.355 & 0.673 / 4.559 & \textcolor{mygreen}{0.591} / \textcolor{mygreen}{3.319} \\
\multirow{5}{*}{} & 1.00 & 0.569 / 2.459 & 0.545 / 2.220 & \textcolor{mygreen}{0.510} / \textcolor{mygreen}{1.753} & 0.543 / 2.346 & 0.665 / 4.460 & 0.766 / 7.179 & \textcolor{mygreen}{0.600} / \textcolor{mygreen}{3.259} & 0.638 / 4.506 \\
\hline
\multirow{4}*{Distance} & 0.5 & 0.608 / 2.914 & 0.589 / 2.738 & \textcolor{mygreen}{0.489} / \textcolor{mygreen}{1.542} & 0.574 / 2.240 & 0.728 / 6.163 & 0.651 / 4.842 & 0.743 / 7.004 & \textcolor{mygreen}{0.564} / \textcolor{mygreen}{3.076} \\
\multirow{4}{*}{} & 1.0 & 0.599 / 2.601 & 0.635 / 3.130 & \textcolor{mygreen}{0.485} / \textcolor{mygreen}{1.609} & 0.571 / 2.529 & 0.802 / 8.173 & 0.681 / 4.460 & 0.737 / 7.455 & \textcolor{mygreen}{0.615} / \textcolor{mygreen}{3.500} \\
\multirow{4}{*}{} & 1.5 & 0.578 / 2.494 & 0.565 / 2.530 & \textcolor{mygreen}{0.523} / \textcolor{mygreen}{1.985} & 0.564 / 2.421 & 0.744 / 6.032 & 0.665 / 4.374 & 0.832 / 8.541 & \textcolor{mygreen}{0.606} / \textcolor{mygreen}{3.324} \\
\multirow{4}{*}{} & 2.0 & 0.594 / 2.824 & 0.589 / 2.952 & \textcolor{mygreen}{0.532} / \textcolor{mygreen}{1.984} & 0.562 / 2.580 & 0.670 / 4.892 & 0.648 / 4.181 & 0.720 / 6.085 & \textcolor{mygreen}{0.589} / \textcolor{mygreen}{3.264} \\
\hline
Region & 1 & 0.652 / 3.088 & 0.757 / 6.973 & 0.646 / \textcolor{mygreen}{2.688} & \textcolor{mygreen}{0.626} / 2.998 & 0.719 / 3.763 & 0.649 / 2.488 & 0.571 / 2.196 & \textcolor{mygreen}{0.490} / \textcolor{mygreen}{1.398} \\ \hline
\end{tabular}
\end{table*}

\begin{table*}[htbp]
\scriptsize
\renewcommand\arraystretch{1.2}
\caption{The attack performance comparison in 3v3 scenarios, measured by two metrics (CR$\downarrow$/CF$\downarrow$).}
\label{tab:add 3v3}
\centering
\begin{tabular}{|c|c|cccc|cccc|}\hline
\multicolumn{2}{|c|}{\multirow{2}*{\textbf{Limitations}}} & \multicolumn{4}{c|}{DDPG} & \multicolumn{4}{c|}{MADDPG}\\ \cline{3-10}
\multicolumn{2}{|c|}{\multirow{2}{*}{}} & \emph{Victim-play} & \emph{APL} & \emph{Victim-curr} & \emph{SUB-PLAY} & \emph{Victim-play} & \emph{APL} & \emph{Victim-curr} & \emph{SUB-PLAY}  \\ \hline
\multirow{5}*{Uncertainty} & 0.00 & 0.652 / 3.393 & 0.676 / 3.146 & 0.652 / 2.944 & \textcolor{mygreen}{0.616} / \textcolor{mygreen}{2.549} & 0.813 / 7.330 & 0.678 / 3.723 & 0.716 / 4.328 & \textcolor{mygreen}{0.644} / \textcolor{mygreen}{3.546} \\
\multirow{5}{*}{} & 0.25 & 0.647 / 2.803 & 0.724 / 3.720 & 0.695 / 3.566 & \textcolor{mygreen}{0.595} / \textcolor{mygreen}{2.685} & 0.682 / 4.958 & 0.751 / 5.466 & 0.770 / 5.094 & \textcolor{mygreen}{0.679} / \textcolor{mygreen}{4.362} \\
\multirow{5}{*}{} & 0.50 & 0.682 / 3.405 & 0.698 / 3.237 & 0.668 / 3.073 & \textcolor{mygreen}{0.647} / \textcolor{mygreen}{3.023} & \textcolor{mygreen}{0.683} / 4.858 & 0.691 / 4.686 & 0.807 / 6.768 & 0.701 / \textcolor{mygreen}{4.386} \\
\multirow{5}{*}{} & 0.75 & 0.684 / 3.260 & 0.693 / 3.127 & 0.724 / 4.514 & \textcolor{mygreen}{0.636} / \textcolor{mygreen}{2.879} & 0.706 / 5.288 & 0.704 / 4.295 & \textcolor{mygreen}{0.685} / \textcolor{mygreen}{3.312} & 0.691 / 4.379 \\
\multirow{5}{*}{} & 1.00 & 0.696 / 3.835 & 0.715 / 3.590 & 0.749 / 4.475 & \textcolor{mygreen}{0.695} / \textcolor{mygreen}{3.330} & 0.684 / 4.552 & 0.812 / 7.709 & \textcolor{mygreen}{0.672} / \textcolor{mygreen}{3.923} & 0.691 / 4.142 \\
\hline
\multirow{4}*{Distance} & 0.5 & 0.670 / \textcolor{mygreen}{3.159} & 0.710 / 3.548 & 0.781 / 5.254 & \textcolor{mygreen}{0.661} / 3.196 & 0.722 / 4.334 & 0.767 / 6.541 & 0.801 / 7.829 & \textcolor{mygreen}{0.665} / \textcolor{mygreen}{3.993} \\
\multirow{4}{*}{} & 1.0 & 0.718 / 3.626 & 0.735 / 3.665 & \textcolor{mygreen}{0.672} / \textcolor{mygreen}{3.140} & 0.705 / 3.189 & 0.761 / 6.329 & 0.700 / 5.285 & \textcolor{mygreen}{0.676} / \textcolor{mygreen}{4.391} & 0.715 / 4.916 \\
\multirow{4}{*}{} & 1.5 & 0.714 / 3.816 & \textcolor{mygreen}{0.687} / \textcolor{mygreen}{3.224} & 0.715 / 3.857 & 0.708 / 3.609 & \textcolor{mygreen}{0.707} / 4.844 & 0.724 / 4.795 & 0.839 / 5.797 & 0.722 / \textcolor{mygreen}{4.227} \\
\multirow{4}{*}{} & 2.0 & 0.720 / 4.381 & 0.672 / \textcolor{mygreen}{3.017} & \textcolor{mygreen}{0.657} / 3.247 & 0.688 / 3.363 & 0.798 / 6.034 & \textcolor{mygreen}{0.691} / \textcolor{mygreen}{4.369} & 0.733 / 4.657 & 0.721 / 4.662 \\
\hline
Region & 1 & 0.829 / 7.350 & 0.950 / 12.130 & 0.768 / 3.896 & \textcolor{mygreen}{0.744} / \textcolor{mygreen}{3.016} & 0.773 / 3.377 & 0.756 / 3.751 & 0.795 / 3.683 & \textcolor{mygreen}{0.696} / \textcolor{mygreen}{2.670} \\ \hline
\end{tabular}
\end{table*}

\begin{table*}[htbp]
\scriptsize
\renewcommand\arraystretch{1.2}
\caption{The attack performance comparison in 2v2 scenarios, measured by two metrics (CR$\downarrow$/CF$\downarrow$).}
\label{tab:add 2v2}
\centering
\begin{tabular}{|c|c|cccc|cccc|}\hline
\multicolumn{2}{|c|}{\multirow{2}*{\textbf{Limitations}}} & \multicolumn{4}{c|}{DDPG} & \multicolumn{4}{c|}{MADDPG}\\ \cline{3-10}
\multicolumn{2}{|c|}{\multirow{2}{*}{}} & \emph{Victim-play} & \emph{APL} & \emph{Victim-curr} & \emph{SUB-PLAY} & \emph{Victim-play} & \emph{APL} & \emph{Victim-curr} & \emph{SUB-PLAY}  \\ \hline
\multirow{5}*{Uncertainty} & 0.00 & 0.466 / 1.733 & 0.410 / 1.121 & 0.444 / 1.409 & \textcolor{mygreen}{0.355} / \textcolor{mygreen}{1.402} & 0.549 / 3.476 & 0.695 / 5.960 & \textcolor{mygreen}{0.361} / \textcolor{mygreen}{1.157} & 0.473 / 2.024 \\
\multirow{5}{*}{} & 0.25 & 0.358 / 1.033 & 0.430 / 1.425 & 0.459 / 2.043 & \textcolor{mygreen}{0.295} / \textcolor{mygreen}{0.826} & 0.567 / 3.712 & 0.624 / 3.958 & \textcolor{mygreen}{0.455} / \textcolor{mygreen}{1.722} & 0.489 / 2.328 \\
\multirow{5}{*}{} & 0.50 & 0.403 / 1.228 & 0.492 / 1.945 & 0.376 / 1.027 & \textcolor{mygreen}{0.351} / \textcolor{mygreen}{0.893} & 0.701 / 5.342 & 0.534 / 2.880 & 0.545 / 2.978 & \textcolor{mygreen}{0.520} / \textcolor{mygreen}{2.767} \\
\multirow{5}{*}{} & 0.75 & 0.369 / 1.162 & 0.396 / 1.037 & 0.382 / 1.016 & \textcolor{mygreen}{0.360} / \textcolor{mygreen}{0.988} & 0.816 / 6.451 & 0.565 / 3.551 & 0.626 / 2.765 & \textcolor{mygreen}{0.468} / \textcolor{mygreen}{1.977} \\
\multirow{5}{*}{} & 1.00 & 0.449 / 1.616 & 0.369 / 1.053 & 0.432 / 1.567 & \textcolor{mygreen}{0.366} / \textcolor{mygreen}{0.949} & 0.810 / 7.168 & 0.500 / 2.597 & 0.698 / 3.223 & \textcolor{mygreen}{0.484} / \textcolor{mygreen}{2.300} \\
\hline
\multirow{4}*{Distance} & 0.5 & 0.409 / 1.285 & \textcolor{mygreen}{0.382} / \textcolor{mygreen}{0.996} & 0.448 / 1.669 & 0.387 / 1.025 & 0.501 / 2.553 & 0.599 / 4.366 & 0.634 / 4.913 & \textcolor{mygreen}{0.463} / \textcolor{mygreen}{2.397} \\
\multirow{4}{*}{} & 1.0 & 0.457 / 1.523 & 0.384 / 1.089 & \textcolor{mygreen}{0.381} / 1.088 & 0.389 / \textcolor{mygreen}{1.061} & 0.509 / 3.211 & 0.576 / 3.501 & 0.492 / 2.662 & \textcolor{mygreen}{0.468} / \textcolor{mygreen}{2.099} \\
\multirow{4}{*}{} & 1.5 & 0.421 / 1.698 & 0.426 / 1.221 & 0.428 / 1.325 & \textcolor{mygreen}{0.389} / \textcolor{mygreen}{1.216} & 0.620 / 3.420 & 0.680 / 4.815 & 0.775 / 6.295 & \textcolor{mygreen}{0.495} / \textcolor{mygreen}{2.291} \\
\multirow{4}{*}{} & 2.0 & 0.428 / 1.580 & 0.412 / 1.349 & \textcolor{mygreen}{0.355} / 1.085 & 0.367 / \textcolor{mygreen}{0.914} & 0.608 / 3.478 & 0.603 / 4.277 & 0.716 / 5.993 & \textcolor{mygreen}{0.485} / \textcolor{mygreen}{2.501} \\
\hline
Region & 1 & 0.604 / 2.677 & \textcolor{mygreen}{0.587} / 2.172 & 0.620 / 2.865 & 0.591 / \textcolor{mygreen}{2.148} & 0.702 / 4.060 & 0.738 / 4.499 & 0.857 / 5.653 & \textcolor{mygreen}{0.489} / \textcolor{mygreen}{1.630} \\ \hline
\end{tabular}
\end{table*}

\begin{table*}[htbp]
\scriptsize
\renewcommand\arraystretch{1.2}
\caption{The attack performance comparison in 4v2 scenarios, measured by two metrics (CR$\downarrow$/CF$\downarrow$).}
\label{tab:add 4v2}
\centering
\begin{tabular}{|c|c|cccc|cccc|}\hline
\multicolumn{2}{|c|}{\multirow{2}*{\textbf{Limitations}}} & \multicolumn{4}{c|}{DDPG} & \multicolumn{4}{c|}{MADDPG}\\ \cline{3-10}
\multicolumn{2}{|c|}{\multirow{2}{*}{}} & \emph{Victim-play} & \emph{APL} & \emph{Victim-curr} & \emph{SUB-PLAY} & \emph{Victim-play} & \emph{APL} & \emph{Victim-curr} & \emph{SUB-PLAY}  \\ \hline
\multirow{5}*{Uncertainty} & 0.00 & 0.580 / 2.062 & 0.603 / 2.156 & 0.604 / \textcolor{mygreen}{1.895} & \textcolor{mygreen}{0.576} / 1.966 & 0.603 / 2.173 & 0.516 / 2.182 & 0.585 / 2.276 & \textcolor{mygreen}{0.509} / \textcolor{mygreen}{1.494} \\
\multirow{5}{*}{} & 0.25 & 0.588 / 1.936 & 0.596 / 2.079 & 0.618 / 2.352 & \textcolor{mygreen}{0.528} / \textcolor{mygreen}{1.572} & 0.604 / 3.345 & 0.571 / 2.473 & 0.557 / 2.037 & \textcolor{mygreen}{0.543} / \textcolor{mygreen}{1.574} \\
\multirow{5}{*}{} & 0.50 & 0.665 / 2.497 & \textcolor{mygreen}{0.559} / 1.910 & 0.583 / \textcolor{mygreen}{1.871} & 0.597 / 2.175 & 0.591 / 2.770 & \textcolor{mygreen}{0.494} / 1.649 & 0.508 / \textcolor{mygreen}{1.574} & 0.555 / 1.864 \\
\multirow{5}{*}{} & 0.75 & 0.616 / 2.127 & 0.654 / 2.298 & 0.683 / 3.812 & \textcolor{mygreen}{0.564} / \textcolor{mygreen}{1.817} & 0.601 / 2.519 & 0.630 / 3.049 & 0.620 / 2.507 & \textcolor{mygreen}{0.544} / \textcolor{mygreen}{1.782} \\
\multirow{5}{*}{} & 1.00 & 0.639 / 2.298 & 0.634 / 2.196 & 0.722 / 3.495 & \textcolor{mygreen}{0.558} / \textcolor{mygreen}{1.767} & 0.643 / 2.583 & 0.546 / 2.369 & 0.615 / 2.877 & \textcolor{mygreen}{0.499} / \textcolor{mygreen}{1.467} \\
\hline
\multirow{4}*{Distance} & 0.5 & 0.672 / 2.578 & \textcolor{mygreen}{0.603} / 2.204 & 0.608 / \textcolor{mygreen}{2.164}  & 0.607 / 2.190 & 0.566 / 2.756 & 0.586 / 2.321 & 0.541 / 1.963 & \textcolor{mygreen}{0.489} / \textcolor{mygreen}{1.554} \\
\multirow{4}{*}{} & 1.0 & 0.653 / 2.419 & 0.625 / 2.433 & 0.616 / 2.215 & \textcolor{mygreen}{0.613} / \textcolor{mygreen}{2.089} & 0.556 / 2.116 & 0.599 / 2.709 & 0.549 / 2.045 & \textcolor{mygreen}{0.528} / \textcolor{mygreen}{1.829} \\
\multirow{4}{*}{} & 1.5 & 0.637 / 2.478 & \textcolor{mygreen}{0.618} / 2.463 & 0.643 / 2.460 & 0.625 / \textcolor{mygreen}{2.256} & 0.656 / 2.960 & 0.579 / 2.307 & 0.570 / 2.378 & \textcolor{mygreen}{0.550} / \textcolor{mygreen}{1.693} \\
\multirow{4}{*}{} & 2.0 & 0.701 / 3.205 & \textcolor{mygreen}{0.574} / \textcolor{mygreen}{1.826} & 0.577 / 1.844 & 0.628 / 2.309 & 0.582 / 2.446 & 0.663 / 2.610 & 0.554 / 2.065 & \textcolor{mygreen}{0.509} / \textcolor{mygreen}{1.592} \\
\hline
Region & 1 & 0.866 / 5.695 & 0.816 / 3.264 & 0.834 / 4.801 & \textcolor{mygreen}{0.791} / \textcolor{mygreen}{2.989} & 0.731 / 2.549 & 0.745 / 2.393 & \textcolor{mygreen}{0.676} / \textcolor{mygreen}{1.847} & 0.721 / 2.366 \\ \hline
\end{tabular}
\end{table*}

\end{document}